\documentclass[conference]{IEEEtran}
\ifCLASSINFOpdf
\else
\fi
\usepackage{amsmath,amssymb,amsfonts}
\usepackage{algorithmic}
\usepackage{graphicx}
\usepackage{textcomp}
\usepackage{xcolor}
\usepackage{url}

\usepackage{amsthm}
\usepackage{amsfonts}
\usepackage{hyperref}
\usepackage{enumerate}
\usepackage{subcaption}
\usepackage{multirow}
\usepackage{placeins}
\usepackage{endnotes}
\usepackage[switch]{lineno}

\theoremstyle{plain}
\newtheorem{theorem}{Theorem}

\newtheorem{lemma}[theorem]{Lemma}

\theoremstyle{definition}

\newtheorem{definition}[theorem]{Definition}

\newtheorem{conjecture}[theorem]{Conjecture}

\begin{document}

%
\title{Optimal $N$-ary ECOC Matrices for Ensemble Classification}


%
\author{\IEEEauthorblockN{Hieu D. Nguyen\IEEEauthorrefmark{1},
Lucas J. Lavalva\IEEEauthorrefmark{2},
Shen-Shyang Ho\IEEEauthorrefmark{2},
Mohammed Sarosh Khan\IEEEauthorrefmark{2} and
Nicholas Kaegi\IEEEauthorrefmark{2}}
\IEEEauthorblockA{\IEEEauthorrefmark{1}Department of Mathematics\\
Rowan University,
Glassboro, NJ 08028\\ Email: nguyen@rowan.edu}
\IEEEauthorblockA{\IEEEauthorrefmark{2}Department of Computer Science\\
Rowan University, Glassboro, NJ 08012\\
Email: lavalv46@students.rowan.edu, hos@rowan.edu, khanmo67@students.rowan.edu, kaegin63@students.rowan.edu}
}

\maketitle

\begin{abstract}
A new recursive construction of $N$-ary error-correcting output code (ECOC) matrices for ensemble classification methods is presented, generalizing the classic doubling construction for binary Hadamard matrices. Given any prime integer $N$, this deterministic construction generates  base-$N$ symmetric square matrices $M$ of prime-power dimension having optimal minimum Hamming distance between any two of its rows and columns.  Experimental results for six datasets demonstrate that using these deterministic coding matrices for $N$-ary ECOC classification yields comparable and in many cases higher accuracy compared to using randomly generated coding matrices.  This is particular true when $N$ is adaptively chosen so that the dimension of $M$ matches closely with the number of classes in a dataset, which reduces the loss in minimum Hamming distance when $M$ is truncated to fit the dataset.  This is verified through a distance formula for $M$ which shows that these adaptive matrices have significantly higher minimum Hamming distance in comparison to randomly generated ones. 
\end{abstract}


%
\IEEEpeerreviewmaketitle

\section{Introduction}

Error correcting output codes (ECOC) is an ensemble machine learning technique introduced by \cite{db} for performing multi-class classfication based on Hamming distance. ECOC is motivated by coding theory where transmitted or stored information is encoded by binary strings (codewords) with high Hamming distance which allows for unique decoding of bit errors.  In particular, each class of a given dataset is assigned a codeword and a predictive model $L=\{L_1,\ldots, L_c\}$ is trained from labeled training data that consists of an ensemble of $c$ binary base learners $L_i$ constructed from the columns of the corresponding coding matrix $M$ (called an ECOC matrix) whose rows consists of the class codewords. Each column of $M$ defines a bipartition of the dataset by merging classes with the same bit value into two superclasses.  Decoding (classification) is performed by matching the codeword predicted by $L$ with the class codeword nearest in Hamming distance. In essence, the ECOC approach is a generalization of one-vs-one and one-vs-all classification techniques.

Hadamard matrices, and in particular Walsh matrices, have been employed as ECOC matrices \cite{gs} since they are optimal in the sense that they have maximum Hamming distance (equal to half the matrix dimension) between any two of its rows and any two of its columns due to symmetry.  Moreover, they can be easily generated recursively using Sylvester's doubling construction and thus only generates square matrices whose dimenions are powers of 2.  To obtain a ECOC matrix whose dimensions fall outside of this case, one typically constructs a Hadamard matrix of larger dimension and then truncate it by deleting a sufficient number of rows and columns to obtain the desired dimension.

The ECOC approach was first generalized to the ternary setting by \cite{allwein} where a third symbol was introduced to represent classes that were removed from training in order to define a sparse encoding. ECOC was then generalized to the $N$-ary setting  by \cite{zhou-wang} where binary codewords (generated from the alphabet $\{0,1\}$) are replaced by codewords generated from the alphabet $\{1,\ldots, N\}$ for any integer $N\geq 2$.  Each column of the corresponding $N$-ary ECOC matrix defines a partition of the set of classes into $N$ superclasses that are trained on an $N$-ary base learner.

 Higher classification accuracies were reported using randomly generated $N$-ary ECOC matrices over binary matrices by \cite{zhou2019} (2019).  Intuitively, this is to be expected since a larger base allows for the existence of ECOC matrices whose rows (and columns) have larger Hamming distance.  The trade-off is that one must train $N$-ary learners as opposed to binary learners, which increases the computational cost.

 In this paper we describe a new recursive construction of $N$-ary ECOC matrices that generalizes Sylvester's doubling construction for Hadamard matrices and generates symmetric square matrices $M$ of dimension $n=N^k$ where for any prime integer $N$.  The Hamming distance for any two rows of $M$ is the same and equals $[(N-1)/N]n$ (a proof is provided in the Appendix).  This formula appears to be optimal with respect to total Hamming distance (defined in Section 3) based on numerical testing.  In addition, we present experimental results that show our deterministic ECOC approach of using these optimal $N$-ary coding matrices (and truncating them to match the dimensions of the desired ECOC matrix) yields classification accuracies that are comparable, and in fact higher for certain datasets where the number of classes is approximately equal to a prime power, to those obtained  where ECOC matrices were found by \cite{zhou-wang} using a best-of-1000 random search.  Therefore, our new construction allows for an efficient and effective design of $N$-ary ECOC matrices.
 
 \section{Related Work}
 
 There are few works that investigate deterministic constructions of data-independent binary ECOC matrices for classification.
 When \cite{db} first introduced ECOC they described four approaches to construct such matrices: using exhaustive codes, selecting good codewords from exhaustive codes, a randomized hill-climbing algorithm, and BCH codes.  Hadamard matrices were later employed as ECOC matrices by \cite{gs} who discussed although an earlier application of Hadamard matrices in neural networks can be found in the work of \cite{chieuh1988}.  A more recent application of Hadamdard ECOC matrices to continual learning was described by \cite{cheng2019} where their model is able to recognize new features without retraining.
 
 For data-independent random ECOC matrices,  \cite{allwein, escalera2009, escalera2010b} investigated various designs, including dense random matrices (binary) and  sparse matrices (ternary), and also described various decoding strategies.   Problem-dependent coding designs have been investigated by \cite{ pujol2006, pujol2008, escalera2010, zhou-wang} (2012).
 
  The approach of using $N$-ary ECOC matrices has been shown by \cite{zhou2019} (2019)  to improve classification accuracy in comparison to binary ECOC.  However, the $N$-ary ECOC matrices used in their work were randomly generated where the best $n\times c$ matrix  was chosen from a batch of 1000, which is rather \textit{inefficient}, especially when their dimensions are quite large.  Our work offers a new deterministic construction that extends Sylvester's recursive construction of Hadamard matrices to the $N$-ary setting.  
  
\section{ECOC Matrices}

Given two binary codewords $\mathbf{x}=(x_0,\ldots, x_{n-1})$ and $\mathbf{y}=(y_0,\ldots, y_{n-1})$, we define their Hamming distance by
\begin{equation}
d_H(\mathbf{x},\mathbf{y})=\sum_{i=0}^{n-1} |x_i-y_i|.
\end{equation}
Let $C=\{\mathbf{x}_0,\ldots,\mathbf{x}_{c-1}\}$ denote an error-correcting output codebook consisting of $c$ binary codewords of length $n$ that encode the $c$ classes of a given dataset. Define $M$ to be the corresponding \textbf{ECOC matrix} of dimension $c\times n$ whose rows are given by the codewords in $C$.  Let $\mathbf{y}_0,\ldots, \mathbf{y}_{n-1}$ denote the $n$ columns of $M$ where each $\mathbf{y}_i$ represents a binary learner that prescribes a partition of the set of classes into two super-classes.  

For machine learning, the following properties are desirable for ECOC matrices \cite{db}:

\begin{itemize}
\item[P1.] Maximum Hamming distance between \textbf{row} codewords $\mathbf{x}_i$ (corresponds to maximum separation between classes).
\item[P2.] Maximum Hamming distance between \textbf{column} codewords $\mathbf{y}_i$ (corresponds to maximum separation between classifiers).
\item[P3.] No \textbf{complements} among column codewords (avoids redundant learners)
\item[P4.] No constant column codewords, i.e., those whose entries of all 0's or all 1's  (avoids grouping all classes into one superclass).
\item[P5.] Ratio $n/c$ that maximizes classification accuracy for a given dataset.
\end{itemize}
In this paper, we focus on properties P1-P4, which are data independent.  Property P5 is data-dependent, although it has been suggested by \cite{allwein} as a heuristic that $n=\lceil 10\log_2{c}\rceil$ be used.  On the other hand, for $N$-aray ECOC, results obtained by \cite{zhou2019} (2019) show that classification accuracy increases as $n$ increases. However, the ECOC matrices used in both works were randomly generated and no comparisons were made with respect to their minimum Hamming distances (between any two rows or columns) as it relates to properties P1-P4. 

To mathematically formulate properties P1-P4, we first introduce notation.  Let $M$ be an $c\times n$ binary matrix consisting of entries in $\{0,1\}$. 

\begin{definition}
Let $d_r(M)$ denote the minimum Hamming distance between any two distinct rows $M$ (called the \textbf{row distance}) and $d_c(M)$ denote the minimum Hamming distance between any two distinct columns of $M$ (called the \textbf{column distance}), i.e., 
\begin{align} 
d_r(M) = & \underset{0\leq i\neq j\leq c-1}{\mathrm{min}} \ d_H(\mathbf{x}_i,\mathbf{x_j}) \label{eq:dr}  \\
d_c(M) = & \underset{0\leq i\neq j \leq n-1}{\mathrm{min}} \ d_H(\mathbf{y}_i,\mathbf{y_j}), \label{eq:dc} 
\end{align}
where $d_H$ denotes the Hamming distance function.  Moreover, we define the \textbf{total distance} of $M$ to be 
\begin{equation} \label{eq:dM} 
d_T(M)=d_r+d_c.
\end{equation}
\end{definition}

\begin{definition} A binary matrix $M$ of dimension $c\times n$ is said to be \textbf{optimal} if it has optimal total Hamming distance, i.e., $d_T(M)$ is maximal over all $c\times n$ binary matrices.  In addition, if $M$ satisfies properties P3 and P4, then it is said to be an \textbf{optimal ECOC} matrix.
\end{definition}

For square dimensions ($n=c$), special Hadamard matrices, called Walsh matrices, yield ECOC matrices with $d_r=d_c=n/2$ so that $d_T(M)=n$.  Walsh matrices can be generated by an efficient recursive construction due to Sylvester.  It is open problem whether Hadamard matrices are optimal for $n\geq 4$ (there are exceptions for $n=3$).  Otherwise, for $m\neq n$, one must resort to exhaustive search to find optimal matrices and it is unclear how optimality is achieved.
 
\subsection{Optimal ECOC Matrices}

Numerical testing suggests the following conjecture regarding an upper bound on optimal binary ECOC matrices in terms of total Hamming distance:

\begin{conjecture} \label{conjecture:bound}
For any square $n\times n$ binary matrix $M$ with $n\geq 3$, we have
\[
d_T(M)\leq \begin{cases}
n, & \textrm{if } n \textrm{ even}; \\
n+1, & \textrm{if } n \textrm{ odd}.
\end{cases}
\]
\end{conjecture}

For $n$ even, and in particular $n=2^k$, the following recursive construction, due to Sylvester, allows for the construction of special Hadamard matrices called Walsh matrices that achieves equality in Conjecture \ref{conjecture:bound}, namely $d_T(M)=n$. 
\begin{lemma} \label{lemma:hadamard}
\cite{gs}
Let $H_k$ be a sequence of Hadamard matrices of dimension $n_k=2^k$ constructed recursively by the block form
\begin{equation}
H_k=\begin{bmatrix}
H_{k-1} & H_{k-1} \\
H_{k-1} & \bar{H}_{k-1} 
\end{bmatrix}, 
H_1=\begin{bmatrix} 0 & 0 \\ 0 & 1 \end{bmatrix} 
\end{equation}
where $\bar{H}_{k-1}$ denotes the binary complement of $H_{k-1}$.  Then the Hamming distance between any two rows (or columns) of $H_k$ equals $n_k/2$.  Thus, $d_T(H_k)=n_k$.
\end{lemma}

For $n=2^k-1$ (odd), equality in Conjecture \ref{conjecture:bound} is achieved  by ``puncturing" the Walsh matrices. 
\begin{lemma}
Defined $P_k$ to be the \textit{punctured} Walsh matrix of dimension $n_k-1$ obtained by deleting the first row and first column of $H_k$. Then the Hamming distance between any two rows (or columns) of $P_k$ equals $n_k/2$.  Thus, $d_T(P_k)=n_k+1$.
\end{lemma}

\section{\textit{N}-ary ECOC Matrices}

In this section we consider $N$-ary ECOC matrices $M=(m_{ij})$ where the entries $m_{ij} \in \{0,1,\ldots, N-1\}$.  There are two approaches to defining Hamming distance for $N$-ary codewords.

\begin{definition}. Let $\mathbf{x}$ and $\mathbf{y}$ be two $N$-ary codewords.
\begin{itemize}
    \item 
   Kronecker delta:  The $N$-ary Hamming distance between $\mathbf{x}$ and $\mathbf{y}$ is defined to be
    \begin{equation} \label{de:kronecker}
        d_H(\mathbf{x},\mathbf{y})=\sum_{i=0}^{n-1} (1- \delta(x_i,y_i)),
    \end{equation}
    where $\delta(x_i,y_i)$ is the Kronecker delta function:
    \[
    \delta(x_i,y_i) = 
    \begin{cases}
    0 & x_i=y_i; \\
    1 & \textrm{otherwise}.
    \end{cases}
    \]
    \item Absolute value: The $N$-ary \textit{absolute} Hamming distance between $\mathbf{x}$ and $\mathbf{y}$ is defined to by
    \begin{equation}\label{de:absolute}
        d_H^{(a)}(\mathbf{x},\mathbf{y})=\sum_{i=0}^{n-1} |x_i-y_i|.
    \end{equation}
\end{itemize}
\end{definition}
Our mathematical results for optimal $N$-ary ECOC matrices described in the next section are based on the Kronecker delta definition given by (\ref{de:kronecker}).  However,  \cite{zhou2019} (2019) used the absolute value definition given by (\ref{de:absolute}) to randomly generate and select the best ECOC matrix.  Experimental results showed no significant difference in accuracy when using either distance functions to randomly search for best-of-1000 ECOC matrices (included in the Appendix). Thus, we report results based only on formula \ref{de:kronecker}. 

Guided by numerical testing and the fact that Hadamard matrices are optimal if we assume Conjecture \ref{conjecture:bound} to be true, we generalize our notion of optimal ECOC binary matrices defined by properties P1-P4 to $N$-ary matrices.  For P3, this requires extending the definition of the complement of a binary codeword to the $N$-ary setting, which can be viewed as a permutation of the labels of the $N$ superclasses.

\begin{definition}\label{de:nary-complement}
Two $N$-ary codewords $\mathbf{x}=(x_0,\ldots,x_{n-1})$ and $\mathbf{y}=(y_0,\ldots,y_{n-1})$ of length $n$ are said to be $N$-ary \textbf{complements} of each other (or complements for short) if there exists a \textit{nontrivial} permutation $f:\{0,\ldots, N-1\} \rightarrow \{0,\ldots, N-1\}$ such that $f(x_i)=y_i$ for all $i\in \{0,\ldots, n-1\}$ and $f(x_i)\neq x_i$ for some $i$.
\end{definition}

Recall our definitions of $d_r(M)$, $d_c(M)$, and $d_T(M)$ given by (\ref{eq:dr}), (\ref{eq:dc}), and (\ref{eq:dM}), respectively, which naturally extend to $N$-ary matrices.  

\begin{definition} \label{de:nary-optimal} 
A square $N$-ary matrix $M$ of dimension $n=N^k$ ($k\in \mathbb{N}$) is said to be \textbf{optimal} if 
$$d_T(M)=2\left(\frac{N-1}{N}\right)n$$ 
over all square $N$-ary matrices of dimension $n$.  In addition, if $M$ satisfies properties P3 and P4 (using Definition \ref{de:nary-complement} above), then it is said to be an \textbf{optimal ECOC} matrix.
\end{definition}

We now present our construction of a family of ECOC optimal $N$-ary square matrices.

\begin{definition} \label{de:nary-ecoc}
We define the square matrix $M_1(N)$ of dimension $N$ whose entries are residues (modulo $N$) obtained by enumerating the values $\{0,1,2,\ldots, N-1\}$ along its diagonals and repeating them beginning with the main diagonal as follows:
\begin{equation*}
M_1(N)=
\begin{bmatrix}
0 & 0 & N-1 & \ldots  & * & * \\
0 & 1 & 1 & \ldots  & * & * \\
N-1 & 1 & 2 & \ldots & * & * \\
\ldots & \ldots & \ldots & \ldots & \ldots & \ldots \\
* & * & * & \ldots & N-2 & N-2 \\
* & * & * & \ldots & N-2 & N-1
\end{bmatrix}
\end{equation*}
or more precisely, if $M_1(N)=[m_{ij}]$, $0\leq i,j \leq N-1$, then its entries $m_{ij}$ are defined by
\begin{equation}\label{eq:shift}
m_{ij} = 
\begin{cases} \displaystyle
i+\sum_{l=1}^{j-i}(N-l+1) \mod N & \textrm {if } i\leq j; \\ \displaystyle
j+\sum_{l=1}^{i-j}(N-l+1) \mod N & \textrm {if } i>j. 
\end{cases}
\end{equation}
\end{definition}
Observe that $M_1(N)$ is symmetric by definition.  Here is the explicit form of $M_1(3)$:
$$
M_1(3)=
\begin{bmatrix}
0 & 0 & 2 \\
0 & 1 & 1 \\
2 & 1 & 2
\end{bmatrix}.
$$

The key idea behind generalizing Sylvester's construction to the $N$-ary setting is to replace the binary complement of a matrix with a modular shift of its entries. 
\begin{definition}\label{de:shift}
Given an $N$-ary matrix $M=[m_{ij}]$ we define a shifted version of it, called the $s$-\textbf{shift} of $M$ and denoted by
$M^{(s)}=[m_{ij}^{(s)}]$, that is
obtained by adding $s$ to each entry of $M$ (modulo $N$), i.e.,
$$
m_{ij}^{(s)}=m_{ij}+s \mod N
$$
\end{definition}

\begin{definition}\label{de:recursive}
We define a sequence of $N$-ary square matrices $M_k:=M_k(N)$ of dimension $n_k=N^k$ recursively by the following block form:
\begin{align*}
M_1 & = M_1(N) \ (\textrm{see Definition } \ref{de:nary-ecoc}) \\
M_{k+1} & = [M_k^{(s_{ij})}], \ \ 0\leq i,j \leq N-1
\end{align*}
where the block $M_k^{(s_{ij})}$ at block row $i$ and block column $j$ denotes the $s_{ij}$-shift of $M_k$ and $s_{ij}=m_{ij}$ is computed from formula (\ref{eq:shift}).
\end{definition}

For example, here is the explicit form for $M_2(3)$:
\begin{align*}
M_2(3) & =
\begin{bmatrix}
0 & 0 & 2 & 0 & 0 & 2 & 2 & 2 & 1 \\
0 & 1 & 1 & 0 & 1 & 1 & 2 & 0 & 0 \\
2 & 1 & 2 & 2 & 1 & 2 & 1 & 0 & 1 \\
0 & 0 & 2 & 1 & 1 & 0 & 1 & 1 & 0 \\
0 & 1 & 1 & 1 & 2 & 2 & 1 & 2 & 2 \\
2 & 1 & 2 & 0 & 2 & 0 & 0 & 2 & 0 \\
2 & 2 & 1 & 1 & 1 & 0 & 2 & 2 & 1 \\
2 & 0 & 0 & 1 & 2 & 2 & 2 & 0 & 0 \\
1 & 0 & 1 & 0 & 2 & 0 & 1 & 0 & 1
\end{bmatrix}    
\end{align*}

The following theorem shows that for prime integers $N$ the matrices $M_k(N)$ are ECOC optimal in the sense of Definition \ref{de:nary-optimal}.

\begin{theorem}\label{th:nary-ecoc}
Let $N\geq 3$ be a prime integer.  Then each $M_k$ is an optimal ECOC matrix.  In particular,
\begin{equation} \label{eq:row-optimal}
d_r(M_k)=d_c(M_k)=\left(\frac{N-1}{N}\right) n_k,
\end{equation}
and thus
\begin{equation} \label{eq:total-optimal}
d_T(M_k)=2\left(\frac{N-1}{N}\right)n_k.
\end{equation}
Moreover, $M_k$ satisfies properties P3 and P4.
\end{theorem}
A proof of Theorem \ref{th:nary-ecoc} is provided in the Appendix.  We note that formula (\ref{eq:total-optimal}) fails for composite integers; for example when $N=4$ and $k=2$, we find that $d_T(M_k)=16$, which according to formula (\ref{eq:total-optimal}) should equal 24; thus, $M_k$ is not optimal.

\section{Experimental Results}

We performed ECOC classification on six public datasets given in Table \ref{tab:dataset}, each having at least 10 classes, to compare classification accuracy between two  strategies:
\begin{enumerate}
    \item Deterministic $N$-ECOC ($N$-ECOC Det): In this strategy each ECOC matrix $M$ was derived from $M_k(N)$ as constructed in Definition \ref{de:recursive}.  For each base $N$, the value $k$ was chosen be the smallest integer for which $n_k$ (the dimension of $M_k(N)$) is larger than or equal to $c$ (number of classes).  We then truncated an appropriate number of rows and columns from $M_k(N)$ (starting from the top left) to obtain our matrix $M$ with dimension $c\times n$. If $M$ happens to contain two identical rows (same codeword for two classes), which occurred only for binary ECOC matrices ($N=2$) and only for dimensions $10\times 5$ (used for Pendigits and Usps) and $11\times 5$ (use for Vowel), then the first entry in one row was flipped to make the codewords be different. 
    
    \item Random $N$-ECOC ($N$-ECOC Rand): In this strategy each ECOC matrix $M_R$ of dimension $n\times c$ was chosen from a batch of 1000 randomly generated matrices and selected to have the largest total distance $d_T(M_R)$. We also considered selecting $M_R$ by optimizing row distance $d_r(M_R)$ (instead of total distance) but found no significant difference in $d_T(M_R)$ and classification accuracy (results provided in the Appendix).  Thus, we only report results for $M_R$ optimized using total distance.

\end{enumerate}

\begin{table}[htbp]
\begin{center}
\begin{tabular}{|c|c|c|c|c|} \hline
Dataset & \# Samples & \# Features & \# Classes \\
\hline 
Pendigits & 3498  & 16 & 10 \\
Usps & 7291 & 256 & 10 \\
Vowel & 990 & 10 & 11 \\
Letters & 20,000 & 16 & 26 \\
Auslan (HQ) & 2565 & 22 & 95 \\
Aloi & 108,000 & 128 & 1000 \\
        \hline
\end{tabular}
\caption{Datasets \label{tab:dataset}}
\end{center}
\end{table}

For both strategies we varied the following parameters:
\begin{itemize}
\item Base $N$: ECOC matrices were constructed for the prime integers $N=2,3,5,7,11,13$.  Larger values were not considered since \cite{zhou2019} (2019) reported little or no improvement in accuracy when $N\geq 10$. 
\item Codeword length $n$ (number of base learners): ECOC matrices of three different dimensions $c\times n$ were constructed, namely $n=0.5c, c, 2c$ (half, square, double, respectively), except for the aloi dataset where we only considered two lengths ($n=0.5c, c$).  This is due to Aloi having a large number of classes ($c=1000$) where it was not computationally feasible to implement $N$-ary ECOC for $n=2c$. 
\item Classification algorithms: We considered two different classifiers $L$ for our base learners: decision tree (DT) and support vector machine (SVM). Default settings were used for each classifier using the implementations  sklearn.tree.DecisionTreeClassifier and sklearn.svm.SVC, respectively, in Python (version 3.7) utilizing its scikit-learn machine learning library.  
\end{itemize}
Thus, given a dataset, we performed 10th-fold cross validation for each set of parameters $\{N,n,L\}$.  Since we are interested in comparing classification performance, we used accuracy as our evaluation metric; thus, mean accuracy and standard deviation are reported for the 10 folds.  Computations were performed on a standard laptop for Pendigits, Usps, Vowel and Letters and on the Open Science Grid \cite{osg} for Auslan and Alo (details given in Appendix).

\subsection{Results and Discussion}

\subsubsection{Comparison of Total (Hamming) Distance}

Tables \ref{tab:distance-pendigits-main}--\ref{tab:distance-aloi-main} reveal that our deterministic construction of $M_k(N)$ and truncating it yields an ECOC matrix $M$ with higher total Hamming distance in comparison to the random approach, but only for those bases $N$ where the dimension of $M$ is close in value to $c$ (number of classes) where little truncation is needed.  For example, take the Letters dataset in Table \ref{tab:distance-letters-main} where $N=k=3$.  The corresponding matrix $M_3(3)$ has dimension  $27\times 27$, and thus it is only necessary to delete a single row and columnn from it to obtain a $26\times 26$ ECOC matrix $M$ having $d_T(M)=34$, whereas the best random ECOC matrix $M_R$ yields a significantly lower total distance of $d_T(M_R)=24$.  On the other hand, the reverse situation occurs for $N=11$ and $k=2$.  The corresponding matrix $M_2(11)$ has dimension $121\times 121$, which requires us to delete 95 rows and columns to obtain a $26\times 26$ ECOC matrix $M$ with $d_T(M)=30$, whereas the best random ECOC matrix $M_R$ has a higher total distance of $d_T(M_R)=40$.  This is because truncating $M_k(N)$, which originally has optimal total distance, results in a matrix $M$ with a lower total distance that is most likely sub-optimal for its dimension. 

\begin{table}[htbp] 
\begin{center}
    \begin{tabular}{|c|c|c|c|c|c|} 
    \hline
        \multicolumn{2}{|c|}{} & \multicolumn{2}{|c|}{$N$-ECOC Det} & \multicolumn{2}{|c|}{$N$-ECOC Rand} \\
    \hline 
        $N$ & $n_k$ & $d_r(M)$ & $d_T(M)$ & $d_r(M_R)$ & $d_T(M_R)$ \\
        \hline
        2 & 16 & 4 & 8  & 3 & 6 \\
        3 & 27 & 6 & 12 & 5 & 10 \\
        5 & 25 & 5 & 10 & 7 & 12 \\
        7 & 49 & 7 & 14 & 7 & 14 \\
    \hline
    \end{tabular}
\caption{\label{tab:distance-pendigits-main} Pendigits - Hamming Distance of $M$ ($10\times 10$)}
\end{center}
\end{table}

\begin{table}[htbp]
\begin{center}
    \begin{tabular}{|c|c|c|c|c|c|} 
    \hline
        \multicolumn{2}{|c|}{} & \multicolumn{2}{|c|}{$N$-ECOC Det} & \multicolumn{2}{|c|}{$N$-ECOC Rand} \\
    \hline 
        $N$ & $n_k$ & $d_r(M)$ & $d_T(M)$ & $d_r(M_R)$ & $d_T(M_R)$ \\
        \hline
        2 & 32 & 12 & 24 & 7 & 15 \\
        3 & 27 & 17 & 34 & 12 & 24 \\
        5 & 125 & 20 & 40 & 16 & 33 \\
        7 & 49 & 19 & 38 & 18 & 36  \\
        11 & 121 & 15 & 30 & 20 & 40 \\
        13 & 169 & 13 & 26 & 21 & 41 \\
    \hline
    \end{tabular}
\caption{\label{tab:distance-letters-main} Letters - Hamming Distance of $M$ ($26\times 26$)}
\end{center}
\end{table}

\begin{table}[htbp]
\begin{center}
    \begin{tabular}{|c|c|c|c|c|c|} 
    \hline
        \multicolumn{2}{|c|}{} & \multicolumn{2}{|c|}{$N$-ECOC Det} & \multicolumn{2}{|c|}{$N$-ECOC Rand} \\
    \hline 
        $N$ & $n_k$ & $d_r(M)$ & $d_T(M)$ & $d_r(M_R)$ & $d_T(M_R)$ \\
        \hline
        2 & 128 & 46 & 92 & 32 & 65 \\
        3 & 243 & 54 & 108 & 48 & 97 \\
        5 & 125 & 70 & 140 & 63 & 127  \\
        7 & 343 & 49 & 98 & 70 & 140  \\
        11 & 121 & 84 & 168 & 77 & 153  \\
        13 & 169 & 82 & 164 & 79 & 157  \\
    \hline
    \end{tabular}
\caption{ \label{tab:distance-auslan-main} Auslan - Hamming Distance of $M$ ($95\times 95$)}
\end{center}
\end{table}

\begin{table}[htbp]
\begin{center}
    \begin{tabular}{|c|c|c|c|c|c|} 
    \hline
        \multicolumn{2}{|c|}{} & \multicolumn{2}{|c|}{$N$-ECOC Det} & \multicolumn{2}{|c|}{$N$-ECOC Rand} \\
    \hline 
        $N$ & $n_k$ & $d_r(M)$ & $d_T(M)$ & $d_r(M_R)$ & $d_T(M_R)$ \\
        \hline
        2 & 1024 & 496 & 992 & 435 & 864  \\
        3 & 2187 & 514 & 1028 & 601 & 1204  \\
        5 & 3125 & 625 & 1250 & 744 & 1487   \\
        7 & 2401 & 657 & 1314 & 808 & 1615   \\
        11 & 1331 & 879 & 1758 & 870 & 1736   \\
        13 & 2197 & 831 & 1662 & 885 & 1770  \\
    \hline
    \end{tabular}
\caption{\label{tab:distance-aloi-main} Aloi - ECOC matrix $M$ ($1000\times 1000$)}
\end{center}
\end{table}

\

\subsubsection{Comparison of Classification Accuracy}
\

\begin{figure}[htbp]
\centerline{\includegraphics[width=140pt,keepaspectratio]{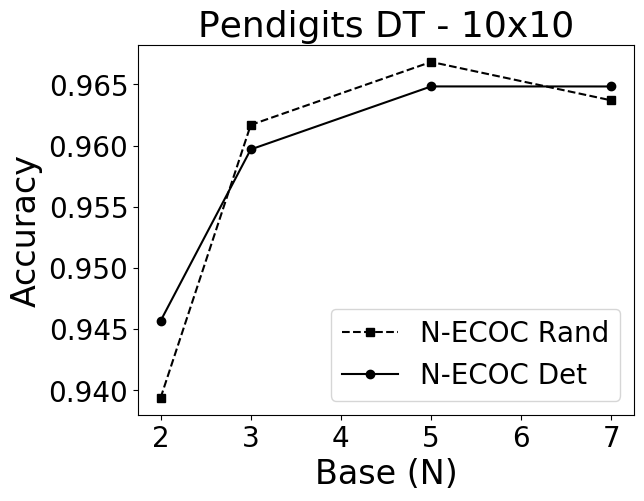}}
\caption{Deterministic and Random $N$-ECOC for Pendigits using DT: Base vs. Accuracy (square)}
\label{fig:pendigits-det-vs-rand-DT-main}
\end{figure}

\begin{figure}[htbp]
\centerline{\includegraphics[width=140pt,keepaspectratio]{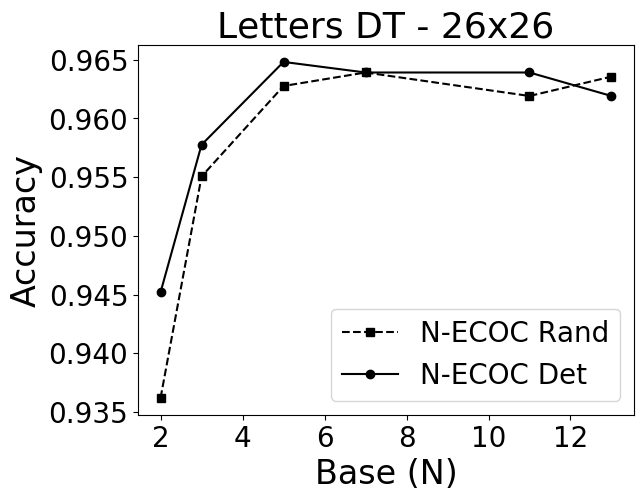}}
\caption{Deterministic and Random $N$-ECOC for Letters using DT: Base vs. Accuracy (square)}
\label{fig:letters-det-vs-rand-DT-main}
\end{figure}

\begin{figure}[htbp]
\centerline{\includegraphics[width=140pt,keepaspectratio]{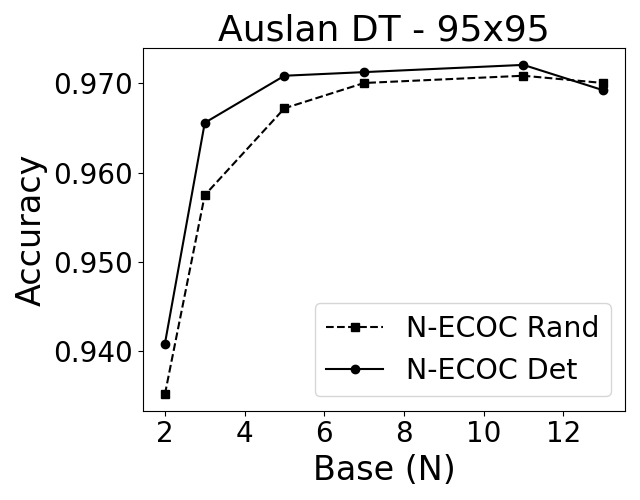}}
\caption{Deterministic and Random $N$-ECOC for Auslan using DT: Base vs. Accuracy (square)}
\label{fig:auslan-det-vs-rand-DT-main}
\end{figure}

\begin{figure}[htbp]
\centerline{\includegraphics[width=140pt,keepaspectratio]{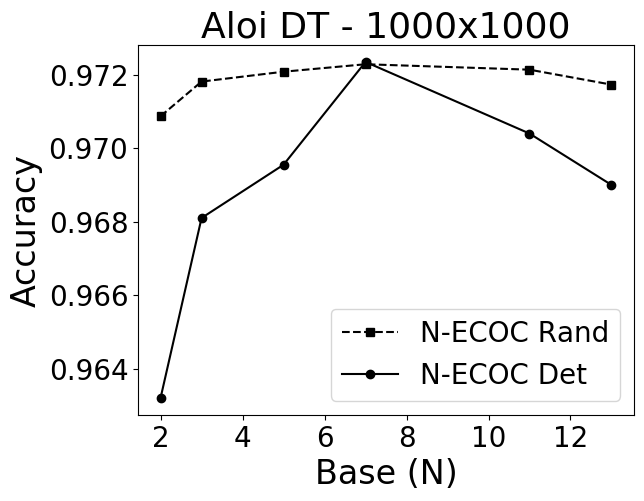}}
\caption{Deterministic and Random $N$-ECOC for Aloi using DT: Base vs. Accuracy (square)}
\label{fig:aloi-det-vs-rand-DT-main}
\end{figure}

\noindent (i) Deterministic vs Random $N$-ECOC:
Figures \ref{fig:pendigits-det-vs-rand-DT-main}-\ref{fig:aloi-det-vs-rand-DT-main} contain plots of classification accuracies for four datasets: Pendigits, Letters, Auslan, and Aloi.  Accuracies were obtained using ECOC matrices of square dimension ($c\times c$) and DT as the classifer for the base learners.  Overall, we found results using our deterministic $N$-ECOC strategy to be competitive with those using the random $N$-ECOC strategy, and in fact higher for those values of $N$ where $n_k$ (dimension of $M_k(N)$) is close in value to $c$.  Recall our earlier discussion in part A involving the Letters dataset, where for the case $N=k=3$ in Table \ref{tab:distance-letters-main}, we saw that the total distance $d_T(M)$ was higher using the deterministic strategy (compared to the random strategy) since little truncation was required, which explains the higher classification accuracy in Figure \ref{fig:letters-det-vs-rand-DT-main}.


\begin{figure}[htbp]
\centerline{\includegraphics[width=140pt,keepaspectratio]{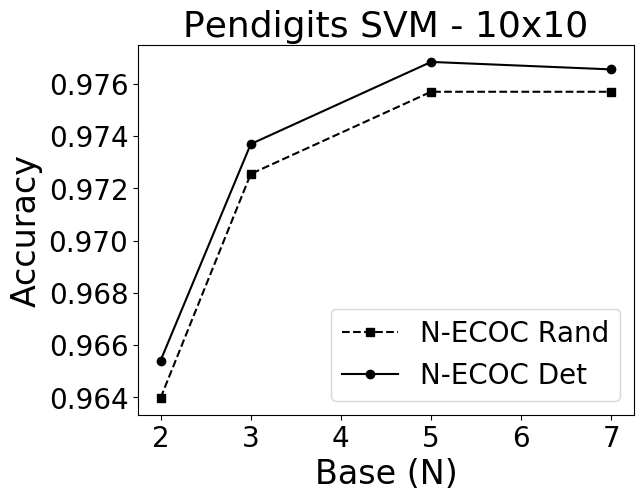}}
\caption{Deterministic and Random $N$-ECOC for Pendigits using SVM: Base vs. Accuracy (square)}
\label{fig:pendigits-det-vs-rand-SVM-main}
\end{figure}

\begin{figure}[htbp]
\centerline{\includegraphics[width=140pt,keepaspectratio]{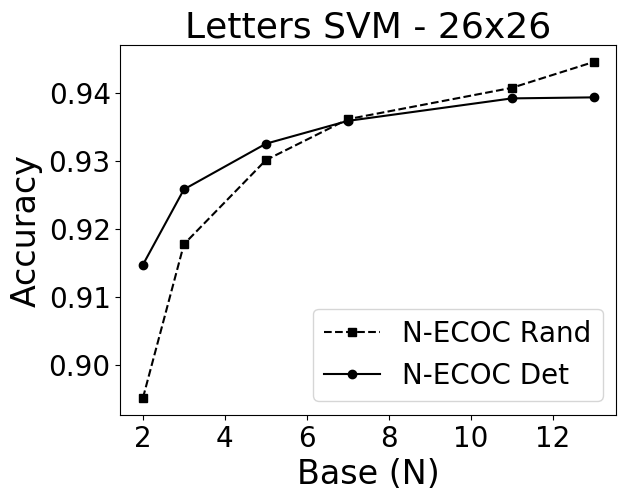}}
\caption{Deterministic and Random $N$-ECOC for Letters using SVM: Base vs. Accuracy (square)}
\label{fig:letters-det-vs-rand-SVM-main}
\end{figure}

\begin{figure}[htbp]
\centerline{\includegraphics[width=140pt,keepaspectratio]{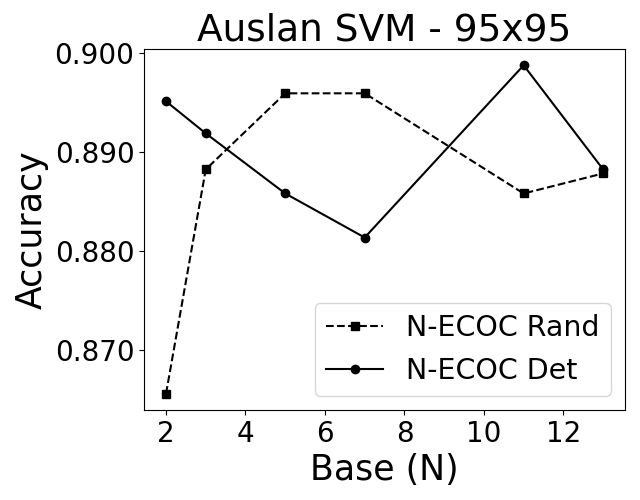}}
\caption{Deterministic and Random $N$-ECOC for Auslan using SVM: Base vs. Accuracy (square)}
\label{fig:auslan-det-vs-rand-SVM-main}
\end{figure}

\begin{figure}[htbp]
\centerline{\includegraphics[width=140pt,keepaspectratio]{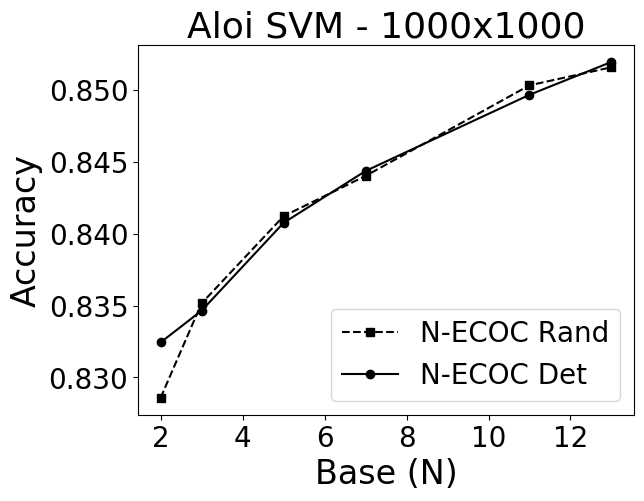}}
\caption{Deterministic and Random $N$-ECOC for Aloi using SVM: Base vs. Accuracy (square)}
\label{fig:aloi-det-vs-rand-SVM-main}
\end{figure}

Thus, it appears that deterministinc $N$-ECOC accuracy is quite competitive with random $N$-ECOC accuracy whenever $n_k\approx c$, a relationshop that holds consistently for all datasets, except for Aloi (see Figure \ref{fig:aloi-det-vs-rand-DT-main} where the random strategy outperformed the determnistic strategy for almost all bases, although the difference is relatively small). We also found this relationship to hold for the classifier SVM; see Figures \ref{fig:pendigits-det-vs-rand-SVM-main}-\ref{fig:aloi-det-vs-rand-SVM-main}.  It is also clear from Figures \ref{fig:pendigits-det-vs-rand-DT-main}-\ref{fig:aloi-det-vs-rand-DT-main} that accuracy increases as the base $N$ increases for both determininistic and random $N$-ECOC, although this increase begins to taper off when $N>7$.

\vspace{5pt}

\begin{figure}
\centerline{\includegraphics[width=140pt,keepaspectratio]{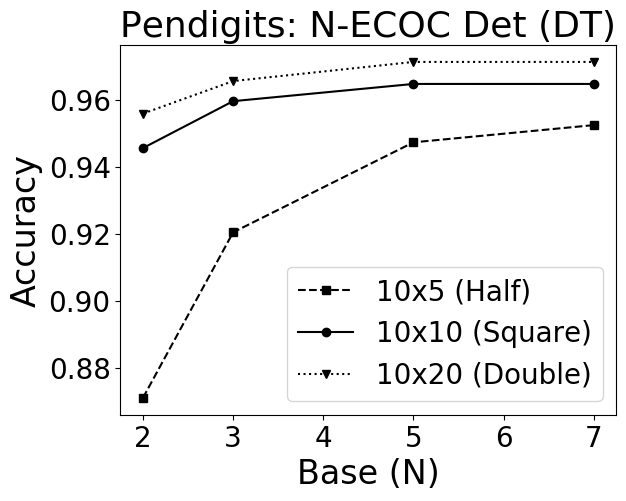}}
\caption{Deterministic $N$-ECOC for Pendigits using Half, Square, and Double dimensions: Base vs. Accuracy (DT)}
\label{fig:pendigits-length-DT-main}
\end{figure}

\begin{figure}
\centerline{\includegraphics[width=140pt,keepaspectratio]{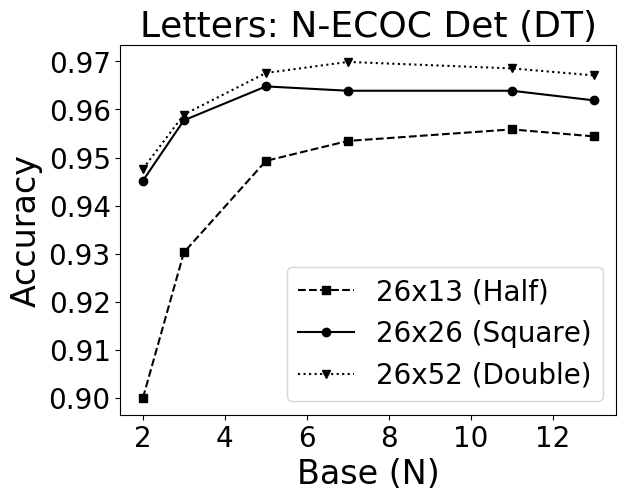}}
\caption{Deterministic $N$-ECOC for Letters using Half, Square, and Double dimensions: Base vs. Accuracy (DT)}
\label{fig:letters-length-DT0-main}
\end{figure}

\begin{figure}
\centerline{\includegraphics[width=140pt,keepaspectratio]{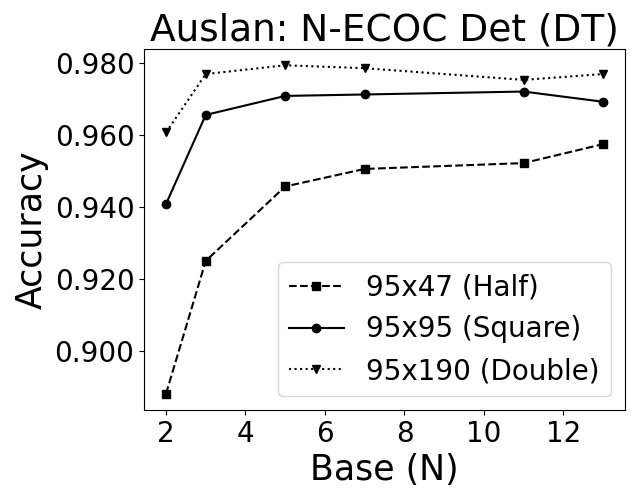}}
\caption{Deterministic $N$-ECOC for Auslan using Half, Square, and Double dimensions: Base vs. Accuracy (DT)}
\label{fig:auslan-length-DT-main}
\end{figure}

\noindent (ii) Varying Matrix Dimension:  When implementing deterministic $N$-ECOC for three different matrix dimensions (half, square, double), we found accuracies in the double case to be consistently higher than the other cases for all datasets, except Aloi.  This is shown in Figures  \ref{fig:pendigits-length-DT-main}-\ref{fig:auslan-length-DT-main} for three datasets (Pendigits, Letters, and Auslan) using DT. Unfortunately, implementing $N$-ECOC for Aloi was not computationally feasible for the double case.  Moreover, we found similar results (provided in the Appendix) to hold when using SVM as classifier.  We also found the double case to be superior for the random $N$-ECOC strategy, which confirms similar results obtained by \cite{zhou2019} (2019) who used random ECOC matrices $M_R$ that were selected by optimizing row distance $d_r(M_R)$ in terms of the absolute Hamming distance function, defined by (\ref{de:absolute}).  There were no significant difference in accuracy in comparison to our random matrices optimized using (\ref{de:kronecker}).


\begin{table*}[htbp]
\begin{center}
    \begin{tabular}{|c|c|c|c|c|} \hline
    & \multicolumn{4}{|c|}{DT} \\
    \hline
 & \multicolumn{2}{|c|}{ECOC (Binary)} & \multicolumn{2}{|c|}{$N$-ECOC (Ternary $N=3$)} \\
\hline
Dataset &  Det & Rand & Det & Rand \\
\hline 
Pendigits & 0.9457 $\pm$ 0.0145 & 0.9394 $\pm$ 0.0082 & 0.9597 $\pm$ 0.0105  & \textbf{0.9617} $\pm$ \textbf{0.0082}  \\
Usps & 0.9056 $\pm$ 0.0132 & 0.8970 $\pm$ 0.0095 & \textbf{0.9365} $\pm$ \textbf{0.0090} & 0.9313 $\pm$ 0.0102 \\
Vowel & 0.8515 $\pm$ 0.0360 & 0.8303 $\pm$ 0.0319 & \textbf{0.9010} $\pm$ \textbf{0.0339} & 0.8919 $\pm$ 0.0320 \\
Letters & 0.9452 $\pm$ 0.0045 & 0.9362 $\pm$ 0.0049 & \textbf{0.9578} $\pm$ \textbf{0.0047} & 0.9551 $\pm$ 0.0035 \\
Auslan & 0.9408 $\pm$ 0.0092 & 0.9352 $\pm$ 0.0117 & \textbf{0.9655} $\pm$ \textbf{0.0092} & 0.9574 $\pm$ 0.0063 \\
Aloi & 0.9632 $\pm$ 0.0044 & 0.9709 $\pm$ 0.0013 & 0.9681 $\pm$ 0.0047 & \textbf{0.9718} $\pm$ \textbf{0.0011} \\
        \hline
    \end{tabular}

    \begin{tabular}{|c|c|c|c|c|} \hline
    & \multicolumn{4}{|c|}{SVM} \\
    \hline
    & \multicolumn{2}{|c|}{ECOC (Binary)} & \multicolumn{2}{|c|}{$N$-ECOC (Ternary $N=3$)} \\
\hline
Dataset &  Det & Rand & Det & Rand \\
\hline 
Pendigits & 0.9654 $\pm$ 0.0111 & 0.9640 $\pm$ 0.0124 & \textbf{0.9737} $\pm$ \textbf{0.0104}  & 0.9725 $\pm$ 0.0082  \\
Usps & 0.9715 $\pm$ 0.0057 & 0.9675 $\pm$ 0.0057 & \textbf{0.9771} $\pm$ \textbf{0.0060} & 0.9740 $\pm$ 0.0062 \\
Vowel & 0.8424 $\pm$ 0.0354 & 0.8101 $\pm$ 0.0398 & 0.8485 $\pm$ 0.0390 & \textbf{0.8576} $\pm$ \textbf{0.0424} \\
Letters & 0.9148 $\pm$ 0.0051 & 0.8952 $\pm$ 0.0069 & \textbf{0.9258} $\pm$ \textbf{0.0061} & 0.9178 $\pm$ 0.0057 \\
Auslan & \textbf{0.8951} $\pm$ \textbf{0.0239} & 0.8655 $\pm$ 0.0177 & 0.8919 $\pm$ 0.0265 & 0.8882 $\pm$ 0.0187 \\
Aloi & 0.8324 $\pm$ 0.0051 & 0.8286 $\pm$ 0.0050 & 0.8346 $\pm$ 0.0048 & \textbf{0.8352} $\pm$ \textbf{0.0044} \\
        \hline
    \end{tabular}
\caption{\label{tab:binary-ternary} Classification accuracy and standard deviation using DT and SVM - binary ECOC vs. ternary $N$-ECOC (square); best performance on each dataset is indicated in bold.}
\end{center}
\end{table*}


\vspace{3pt}
\noindent (iii) ECOC (Binary) vs. $N$-ECOC: It is clear from previous figures that the $N$-ECOC approach is superior to standard ECOC (binary) approach, which shows higher accuracies when $N\geq 3$ compared to $N=2$ (binary), regardless of classifier and matrix dimension.  Accuracy appears to increase as the base increases, but the gain in accuracy diminishes when $N > 7$, which confirms the results of \cite{zhou2019} (2019).  Table \ref{tab:binary-ternary} shows accuracies for binary ECOC versus ternary $N$-ECOC ($N=3$), where the latter show superior performance over the former across many datasets (best performance for each dataset is indicated in bold).


\section{Conclusion}

In summary we have shown our deterministic $N$-ECOC strategy, based on a novel, deterministic, and recursive construction of optimal $N$-ary matrices,  to yield competitive accuracies in comparison to a random $N$-ECOC strategy.  In particular, the latter is quite effective for those data sets where the number of classes is approximately equal to a prime power.  We believe our work will shed light on methods for finding optimal $N$-ary square matrices for non-prime powers, which will be part of future work, as well as adaptive truncation methods to minimize loss of total distance.


\bibliographystyle{IEEEtran}

\bibliography{nguyen_SSCI2021}

\section{Appendix}
This technical appendix contains two parts.  The first part gives a proof of Theorem 12 stated in the main paper.  The second part provides additional experimental results to support claims made in the main paper.

\subsection{Proof of Theorem 12 (Main Paper)}

We first establish several lemmas that will be needed in the proof.  Towards this end, we begin by defining the ``modular" distance between two vectors.

\begin{definition} 
Given two $N$-ary vectors $u=(u_0,\ldots, u_{n-1})$ and $v=(v_0,\ldots, v_{n-1})$, we define their \textbf{modular difference vector} $\Delta(u,v)$ to be
\[
\Delta(u,v)=(\delta_0,\ldots, \delta_{n-1}),
\]
where $\delta_i=(u_i-v_i) \mod N$.  
\end{definition}

Observe that the Hamming distance defined by (6) in the main paper is related to the modular difference vector by
$$d_H(u,v)=|\{i: \delta_i\neq 0\}|.$$
The following lemma reveals the ideas behind our recursive construction of $N$-ary matrices of Hadamard type.

\begin{lemma} \label{le:nary-base-case}
Let $N\geq 3$ be a prime integer and $M_1:=M_1(N)$ be the $N$-ary matrix defined as in Definition 9 of the main paper.  

\begin{enumerate}[(a)]
\item Let $r_i=(m_{i0},\ldots, m_{i(N-1)})$ and $r_j=(m_{j0},\ldots, m_{j(N-1}))$ denote the $i$-th and $j$-th rows of $M_1$, respectively.  If $i=j$, then $\Delta(r_i,r_j)=(0,\ldots,0)$ and thus $d_H(r_i,r_j)=0$.  If $i\neq j$, then $\Delta(r_i,r_j)$ is a permutation of $\{0,1,\ldots, N-1\}$ and thus $d_H(r_i,r_j)=N-1$.  The same result holds for any two rows of $M_1^{(s)}$ for $s\in \{0,1,\ldots, N-1\}$.

\item Let $r_i^{(s)}$ and $r^{(t)}_j$ denote the $i$-th and $j$-th rows of $M_1^{(s)}$ and $M_1^{(t)}$, respectively, for any two non-negative integers $s$ and $t$ with $s\neq t$.  If $i=j$, then $\Delta(r_i^{(s)},r_j^{(t)})=(a,\ldots,a)$ (constant vector) where 
$$a=s-t \mod N$$ 
and thus $d_H(r_i^{(s)},r_j^{(t)})=N$.    If $i\neq j$, then $\Delta(r_i^{(s)},r_j^{(t)})$ is a permutation of $\{0,1,\ldots, N-1\}$, and thus $d_H(r_i^{(s)},r_j^{(t)})=N-1$.
\end{enumerate}

\end{lemma}

\begin{proof}
We first prove part (a) for any two rows $r_i$, $r_j$ of $M_1$.  The result is clear if $i=j$.  Therefore, assume without loss of generality that $i < j$.  We use equation (8) in the main paper to compute $\Delta(r_i,r_j)=(\delta_0,\ldots, \delta_{N-1})$ by considering three cases:

\noindent Case 1: $i < j \leq k$, we have
\begin{align*}
    \delta_k & = (m_{ik} - m_{jk}) \mod N \\
    & = [i+\sum_{l=1}^{k-i}(N-l+1)]  \\
    & \ \ \ \ - [j+\sum_{l=1}^{k-j}(N-l+1)] \mod N \\
    & = i-j - \frac{1}{2}(i - j) (1 + i + j - 2 k) \mod N \\
    & = k(i-j) - \frac{1}{2}(i-j)(i+j-1) \mod N
\end{align*}

\noindent Case 2: $k \leq i < j$, we have
\begin{align*}
    \delta_k & = (m_{ik} - m_{jk}) \mod N \\
    & = [k+\sum_{l=1}^{i-k}(N-l+1)]  \\
    & \ \ \ \ - [k+\sum_{l=1}^{j-k}(N-l+1)] \mod N \\
    & = k(i-j) - \frac{1}{2}(i-j)(i+j-1) \mod N
\end{align*}

\noindent Case 3: $i < k < j$, we have
\begin{align*}
    \delta_k & = (m_{ik} - m_{jk}) \mod N \\
    & = [i+\sum_{l=1}^{k-i}(N-l+1)]  \\
    & \ \ \ \ - [k+\sum_{l=1}^{j-k}(N-l+1)] \mod N \\
    & = k(i-j) - \frac{1}{2}(i-j)(i+j-1) \mod N
\end{align*}
Since the formula for $\delta_k$ is identical in all three cases and is linear in $k$ with $i-j\neq 0$, it follows that $\delta_k$ that takes on all residues for any prime modulus $N$ when $k=0,\ldots, N-1$.  Thus, $\delta_k$ is a permutation of $\{0,1,\ldots, N-1\}$.

For any two rows $r_i^{(s)}$, $r_j^{(s)}$ of $M_1^{(s)}$, we define $\Delta(r_i^{(s)},r_j^{(s)})=(\delta_0^{(s)},\ldots,\delta_{N-1}^{(s)}$.  The the same result now follows easily from the fact that
\begin{align*}
\delta_k^{(s)} & = (m^{(s)}_{ik} - m^{(s)}_{jk}) \mod N \\
& = (m_{ik} - m_{jk}) \mod N \\
& = \delta_k.
\end{align*}

We now prove part (b).  Let $r_i^{(s)}$ and $r_j^{(t)}$ denote the $i$-th and $j$-th rows of $M_1^{(s)}$ and $M_1^{(t)}$, respectively, with $s\neq t$.  We define $\Delta(r_i^{(s)},r_j^{(t)})=(\delta_0^{(ts)},\ldots,\delta_{N-1}^{(st)}$.  If $i=j$, then
\begin{align*}
\delta_k^{(st)} & = (m_{ik}^{(s)} - m_{jk}^{(t)}) \mod N \\
    & = s-t \mod N \\
    & = a
\end{align*}
Thus, $\Delta(r_i^{(s)},r_j^{(j)})=(a,\ldots,a)$ and $d_H(r_i,r'_j)=N$.  

Next, assume with loss of generality that $i<j$.  We have
\begin{align*}
\delta_k^{(st)} 
    & = (m_{ik}^{(s)} - m_{jk}^{(t)}) \mod N \\
    & = s-t + m_{ik}-m_{jk}  \mod N \\
    & = s-t + k(i-j) - \frac{1}{2}(i-j)(i+j-1) \mod N
\end{align*}
Again, since $\delta_k^{(st)}$ is linear in $k$ with $i-j\neq 0$, it follows that $\delta_k^{(st)}$ generates all residues for any prime modulus $N$ when $k=1,\ldots, N$.  Thus, $\Delta(r_i^{(s)},r_j^{(t)})$ is a permutation of $\{0,1,\ldots, N-1\}$ and $d_H(r_i^{(s)},r_j^{(t)})=N-1$.
\end{proof}

\begin{definition} We shall say that two $N$-ary vectors $u$ and $v$ of length $n=Np$ have \textbf{multiplicity} $p$ if $m(u,v)$ is a multi-permutation of $\{0,1,\dots, N-1\}$, i.e., a vector where each $N$-ary symbol occurs $p$ times, and thus $d_H(u,v)=(N-1)p$.  Similarly, we shall say that a matrix $M$ has multiplicity $p$ if any two distinct rows of $M$ have multiplicity $p$.

\end{definition}

It is clear from Lemma \ref{le:nary-base-case} that $M_1(N)^{(s)}$ has multiplicity 1 for any non-negative integer $s$.  The next lemma extends this result to matrices with higher multiplicity, which can be easily proven by using the same arguments as in the proof of Lemma \ref{le:nary-base-case}.  Thus, we omit the proof.

\begin{lemma} \label{le:multiplicityp}
Let $M$ be a $N$-ary matrix of dimension $n=Np$ with multiplicity $p$, i.e., any two distinct rows of $M$ have multiplicity $p$.  Define $M^{(s)}$ as in Definition 10 (main paper).  Then

\begin{enumerate}[(a)]
\item $M^{(s)}$ has multiplicity $p$.

\item Let $r_i^{(s)}$ and $r_j^{(t)}$ denote the $i$-th and $j$-th rows of $M^{(s)}$ and $M^{(t)}$, respectively, with $s\neq t$.  If $i=j$, then $\Delta(r_i^{(s)},r_j^{(t)})=(a,\ldots, a)$ (constant vector) where $a\in \{1,\ldots, N-1\}$ and thus $d_H(r_i,r'_j)=n$.    If $i\neq j$, then $r_i^{(s)}$ and $r_j^{(t)}$ have multiplicity $p$ and thus $d_H(r_i,r'_j)=(N-1)p$.
\end{enumerate}

\end{lemma}

We next establish that $M_1(N)$ satisfies property P3.

\begin{lemma}
\label{le:nary-complement}
Let $N\geq 3$ be a prime integer and $M_1:=M_1(N)$ be the $N$-ary matrix defined as in Definition 9 (main paper).  Then no two rows (or columns) are $N$-ary complements.
\end{lemma}

\begin{proof}
Since $M_1$ is symmetric, it suffices to prove that any two distinct rows of $M_1$, denoted by $r_i=(m_{i0},\ldots, m_{i(n-1)})$ and $r_j=(m_{j0},\ldots, m_{j(n-1)})$, are not $N$-ary complements.  Recall that we proved earlier in Lemma \ref{le:nary-base-case} that  $\delta_k=(m_{ik}-m_{jk}) \mod N$ takes on all residues when $k=0,\ldots, N-1$.  Thus, there exists a unique $k_0$ for which $\delta_{k_0}=0$, i.e., $m_{ik_0}=m_{jk_0}$.  We claim that there exists $k_1$ such that $m_{jk_0}=m_{jk_1}$, but $m_{ik_1}\neq m_{jk1}$.
This proves that no permutation $f:\{0,\ldots, N-1\}\rightarrow \{0,\ldots, N-1\}$ exists with $f(m_{ik})=m_{jk}$, and thus $r_i$ and $r_j$ are \textit{not} $N$-ary complements.

To prove our claim, we assume $i<j$ and $N\geq 5$ (the case $N=3$ can be easily verified by brute force), and use the fact that $m_{kl}$ is periodic in $k$ and $l$ with period $N$:
$$m_{kl}=m_{(k\pm N)l}=m_{k(l\pm N)}$$
Set 
\begin{align*}
k_0' & =N+i + \frac{1}{2}(N + 1)(j - i - 1) \\
k_1' & =N+j + \frac{1}{2}(N + 1)(j - i - 1) + 2 \\
k_0 & =k_0' \mod N \\
k_1 & =k_1' \mod N
\end{align*}
Then
\begin{align*}
    \delta_{k_0}& :=m_{jk_0}-m_{ik_0} \mod N \\
    & = m_{jk_0'}-m_{ik_0'} \mod N \\
    &= \frac{1}{2} (i - j) (i - j + 1)N \mod N \\
    & = 0
\end{align*}
Thus, $m_{ik_0}=m_{jk_0}$.  By the same argument, we have
\begin{align*}
    \gamma & :=m_{jk_0}-m_{jk_1} \mod N \\
    & = m_{jk_0'}-m_{jk_1'} \mod N \\
    &= \frac{1}{2} ((i - j) (i - j - 1) - 2)N \mod N \\
    & = 0
\end{align*}
Thus, $m_{jk_0}=m_{jk_1}$.  On the other hand,
\begin{align*}
    \delta_{k_1}& :=m_{jk_1}-m_{ik_1} \mod N \\
    & = m_{jk_1'}-m_{ik_1'} \mod N \\
    &= \frac{1}{2} (i-j-2) (N (i - j + 1) + 2 (i - j)) \mod N \\
    & = (i-j-2) (i - j) \mod N \\
    & \neq 0
\end{align*}
since $N\geq 5$ is prime.  Thus, $m_{jk_1}\neq m_{ik_1}$.
\end{proof}

The following lemma extends Lemma \ref{le:nary-complement} to shifts of $M_1(N)$.  The proof is straightforward, which we omit.
\begin{lemma}
\label{le:nary-complement-extended}
Let $r_i^{(s)}$ and $r_j^{(t)}$ be the $i$-th and $j$-th rows of $M_1^{(s)}(N)$ and $M_1^{(t)}(N)$, respectively, with $i\neq j$.  Then $r_i^{(s)}$ and $r_j^{(t)}$ are not $N$-ary complements.
\end{lemma}

We are now ready to prove Theorem 12 in the main paper, which we restate to include a result regarding the multiplicity of $M_k(N)$.

\begin{theorem}[Theorem 12 in main paper] \label{thm12}
Let $M_k:=M_k(N)$ be a $N$-ary matrix of dimension $n_k=N^k$, defined recursively as in Definition 11 (main paper).
Then $M_k$ has multiplicity $N^{k-1}$ and
\begin{equation}
    d_H(r_i,r_j)=(N-1)N^{k-1}
\end{equation}
for any two distinct rows $r_i$, $r_j$ of $M_k$.
Thus, $M_k$ has minimum row distance (and minimum column distance) of $((N-1/N))n_k$, i.e.,
\[
d_r(M_k)=d_c(M_k)=(N-1)N^{k-1}=\left(\frac{N-1}{N}\right) n_k.
\]

\end{theorem}

\begin{proof}
We prove by induction that $M_k$ has multiplicity $N^{k-1}$.  It is clear from Lemma \ref{le:nary-base-case} that the result is true for $k=1$, namely $M_1$ has multiplicity 1 and $d_r(M_1)=d_c(M_1)=N-1$.  Next, assume that the result holds for $M_k$, namely that $M_k$ (and in fact every $M_k^{(s)}$) has multiplicity $N^{k-1}$.  We prove that $M_{k+1}$ has multiplicity $N^{k+1}$.
Let $r_i$ and $r_j$ be two distinct rows of $M_{k+1}$.  Write $i=q_iN+p_i$ and $j=q_jN+p_j$.  Using the recursive construction for $M_{k+1}$, we  view $r_i$ as a concatenation (denoted by the $\#$) of $N$ rows, $r_{p_i}^{(s_{q_i1})}, \ldots, r_{p_i}^{(s_{q_iN})}$, where $r_{p_i}^{(s_{q_il})}$ is the $p_i$-th row of $M_k^{(s_{q_il})}$, and similarly for $r_j$:
\begin{align} \label{eq:concat1}
r_i & =r_{p_i}^{(s_{q_i1})}\#r_{p_i}^{(s_{q_i2})}\# \cdots \# r_{p_i}^{(s_{q_iN})} \\ \label{eq:concat2}
r_j & =r_{p_j}^{(s_{q_j1})}\#r_{p_j}^{(s_{q_j2})}\# \cdots \# r_{p_j}^{(s_{q_jN})}
\end{align}

We now consider two cases.  

\noindent CASE I: $p_i=p_j$.  Then since $q_i\neq q_j$ and $M_1$ has multiplicity 1, there exists a unique $l_0\in \{0,1,\ldots, N-1\}$ such that $$s_{q_il_0}=m_{q_1l_0}=m_{q_jl_0}=s_{q_jl_0}$$
and so $\Delta(r_{p_i}^{(s_{q_il_0})},r_{p_j}^{(s_{q_jl_0})})=(0,\ldots, 0)$.  It follows that
$$d_H(r_{p_i}^{(s_{q_il_0})},r_{p_j}^{(s_{q_jl_0})})=0.$$  
For $l\neq l_0$, we have $s_{q_il_0}\neq s_{q_jl_0}$ by the same reason and it follows again from Lemma \ref{le:multiplicityp} that $\Delta(r_{p_i}^{(s_{q_il_0})},r_{p_j}^{(s_{q_jl_0})})=(a,\ldots, a)$ where $a$ takes on each value in $\{1,\ldots, N-1\}$ for each value of $l\neq l_0$.  It follows that
$$
d_H(r_{p_i}^{(s_{q_il})},r_{p_j}^{(s_{q_jl})})=n_k
$$
Thus,
\begin{align*}
d_H(r_i,r_j) & = d_H(r_{p_i}^{(s_{q_i1})},r_{p_j}^{(s_{q_j1})})+ \ldots  \\
& \ \ \ \ + d_H(r_{p_i}^{(s_{q_i(N-1)})},r_{p_j}^{(s_{q_j(N-1)})})  \\
& = (N-1)n_k \\
& = (N-1)N^{k}
\end{align*}
Moreover, because of (\ref{eq:concat1}) and (\ref{eq:concat2}), $r_i$ and $r_j$ have multiplicity $N\cdot N^{k-1}=N^k$.

\noindent CASE II: $p_i\neq p_j$.  By Lemma \ref{le:multiplicityp}, we have that $r_{p_i}^{(s_{q_il})}$ and $r_{p_j}^{(s_{q_jl})}$ have multiplicity $N^{k-1}$.  It follows that
$$
d_H(r_{p_i}^{(s_{q_il})},r_{p_j}^{(s_{q_jl})})=(N-1)N^{k-1}
$$
and thus
\begin{align*}
d_H(r_i,r_j) & = d_H(r_{p_i}^{(s_{q_i1})},r_{p_j}^{(s_{q_j1})})+ \ldots  \\
& \ \ \ \ + d_H(r_{p_i}^{(s_{q_i(N-1)})},r_{p_j}^{(s_{q_j(N-1)})})  \\
& = (N-1)N^{k-1} + \ldots + (N-1)N^{k-1} \\
& = (N-1)N^{k}
\end{align*}
Again, $r_i$ and $r_j$ have multiplicity $N\cdot N^{k-1}=N^k$ by the same reason.

Next, we prove by induction that $M_k$ satisfies property P3.  Since $M_k$ is symmetric, it suffices to prove that no two distinct rows are $N$-ary complements.  For $M_1$, this was established in Lemma \ref{le:nary-complement}.  Suppose then that no two rows of $M_k$ are $N$-ary complements.  We shall prove the same for any two rows $r_i$ and $r_j$ be two rows of $M_{k+1}$.   Recall the decomposition (concatenation) given by (\ref{eq:concat1}) and (\ref{eq:concat2}):
\begin{align}
r_i & =r_{p_i}^{(s_{q_i1})}\#r_{p_i}^{(s_{q_i2})}\# \cdots \# r_{p_i}^{(s_{q_iN})} \\ 
r_j & =r_{p_j}^{(s_{q_j1})}\#r_{p_j}^{(s_{q_j2})}\# \cdots \# r_{p_j}^{(s_{q_jN})}
\end{align}
We consider two cases:

CASE I: $p_i=p_j$.  By the same argument as in Lemma \ref{le:nary-complement-extended}, there exists $k_0$ and $k_1$ for which (recall the notation used in Definition 11 in the main paper)
\begin{align*}
    s_{q_ik_0} & =m_{q_ik_0}=m_{q_jk_0}=s_{q_jk_0} \\
    s_{q_jk_0} & =m_{q_jk_0}=m_{q_jk_1}=s_{q_jk_1} \\
    s_{q_ik_1} & =m_{q_ik_1}\neq m_{q_jk_1}=s_{q_jk_1}
\end{align*}
It follows that
\begin{align*}
    r_{p_i}^{(s_{q_ik_0})} & = r_{p_j}^{(s_{q_jk_0})} \\
    r_{p_j}^{(s_{q_jk_0})} & = r_{p_j}^{(s_{q_jk_1})} \\
    r_{p_i}^{(s_{q_ik_1})} & \neq r_{p_j}^{(s_{q_jk_1})}
\end{align*}
Thus, $r_i$ and $r_j$ are not $N$-ary complements.

CASE II: $p_i\neq p_j$.  It follows by Lemma \ref{le:nary-complement-extended} that $r_{p_i}^{(s_{q_i1})}$ and $r_{p_j}^{(s_{q_j1})}$ are not $N$-ary complements.  Thus, $r_i$ and $r_j$ are not $N$-ary complements.

Lastly, it is clear by the recursive definition of $M_k(N)$ that it satisfies property P4, namely none of its columns are constant codwords, i.e., all entries are not the same.  This completes the proof of  Theorem \ref{thm12}.
\end{proof}

\subsection{Additional Experimental Results}

In this section we provide additional experimental results to support claims made in the main paper.

\subsubsection{Comparing Minimum Distances of ECOC matrices for Deterministic vs Random $N$-ECOC (Hamming Distance)}

We provide additional results beyond those presented in the main paper by reporting row distances $d_r(M)$ and $d_r(M_R)$, and total distances $d_T(M)$ and $d_T(M_R)$ for all datasets, where $M$ and $M_R$ are square ECOC matrices generated deterministically and randomly as discussed in the main paper. 

\begin{table}[htbp] 
\begin{center}
    \begin{tabular}{|c|c|c|c|c|c|} 
    \hline
        \multicolumn{2}{|c|}{} & \multicolumn{2}{|c|}{$N$-ECOC Det} & \multicolumn{2}{|c|}{$N$-ECOC Rand} \\
    \hline 
        $N$ & $n_k$ & $d_r(M)$ & $d_T(M)$ & $d_r(M_R)$ & $d_T(M_R)$ \\
        \hline
        2 & 16 & 4 & 8  & 3 & 6 \\
        3 & 27 & 6 & 12 & 5 & 10 \\
        5 & 25 & 5 & 10 & 7 & 12 \\
        7 & 49 & 7 & 14 & 7 & 14 \\
    \hline
    \end{tabular}
\caption{\label{tab:distance-pendigits} Pendigits - Hamming Distance of $M$ ($10\times 10$)}
\end{center}
\end{table}

\begin{table}[htbp] 
\begin{center}
    \begin{tabular}{|c|c|c|c|c|c|} 
    \hline
        \multicolumn{2}{|c|}{} & \multicolumn{2}{|c|}{$N$-ECOC Det} & \multicolumn{2}{|c|}{$N$-ECOC Rand} \\
    \hline 
        $N$ & $n_k$ & $d_r(M)$ & $d_T(M)$ & $d_r(M_R)$ & $d_T(M_R)$ \\
        \hline
        2 & 16 & 4 & 8 & 3 & 6 \\
        3 & 27 & 6 & 12 & 5 & 10 \\
        5 & 25 & 5 & 10 & 7 & 13 \\
        7 & 49 & 7 & 14 & 7 & 14 \\
    \hline
    \end{tabular}
\caption{\label{tab:distance-usps} Usps - Hamming Distance of $M$ ($10\times 10$)}
\end{center}
\end{table}

\begin{table}[htbp] 
\begin{center}
    \begin{tabular}{|c|c|c|c|c|c|} 
    \hline
        \multicolumn{2}{|c|}{} & \multicolumn{2}{|c|}{$N$-ECOC Det} & \multicolumn{2}{|c|}{$N$-ECOC Rand} \\
    \hline 
        $N$ & $n_k$ & $d_r(M)$ & $d_T(M)$ & $d_r(M_R)$ & $d_T(M_R)$ \\
        \hline
        2 & 16 & 4 & 8 & 4 & 7 \\
        3 & 27 & 6 & 12 & 5 & 10 \\
        5 & 25 & 6 & 12 & 7 & 14\\
        7 & 49 & 7 & 14 & 8 & 16 \\
        11 & 11 & 10 & 20 & 9 & 17 \\
    \hline
    \end{tabular}
\caption{\label{tab:distance-vowel} Vowel - Hamming Distance of $M$ ($11\times 11$)}
\end{center}
\end{table}

\begin{table}[htbp]
\begin{center}
    \begin{tabular}{|c|c|c|c|c|c|} 
    \hline
        \multicolumn{2}{|c|}{} & \multicolumn{2}{|c|}{$N$-ECOC Det} & \multicolumn{2}{|c|}{$N$-ECOC Rand} \\
    \hline 
        $N$ & $n_k$ & $d_r(M)$ & $d_T(M)$ & $d_r(M_R)$ & $d_T(M_R)$ \\
        \hline
        2 & 32 & 12 & 24 & 7 & 15 \\
        3 & 27 & 17 & 34 & 12 & 24 \\
        5 & 125 & 20 & 40 & 16 & 33 \\
        7 & 49 & 19 & 38 & 18 & 36  \\
        11 & 121 & 15 & 30 & 20 & 40 \\
        13 & 169 & 13 & 26 & 21 & 41 \\
    \hline
    \end{tabular}
\caption{\label{tab:distance-letters} Letters - Hamming Distance of $M$ ($26\times 26$)}
\end{center}
\end{table}

\begin{table}[htbp]
\begin{center}
    \begin{tabular}{|c|c|c|c|c|c|} 
    \hline
        \multicolumn{2}{|c|}{} & \multicolumn{2}{|c|}{$N$-ECOC Det} & \multicolumn{2}{|c|}{$N$-ECOC Rand} \\
    \hline 
        $N$ & $n_k$ & $d_r(M)$ & $d_T(M)$ & $d_r(M_R)$ & $d_T(M_R)$ \\
        \hline
        2 & 128 & 46 & 92 & 32 & 65 \\
        3 & 243 & 54 & 108 & 48 & 97 \\
        5 & 125 & 70 & 140 & 63 & 127  \\
        7 & 343 & 49 & 98 & 70 & 140  \\
        11 & 121 & 84 & 168 & 77 & 153  \\
        13 & 169 & 82 & 164 & 79 & 157  \\
    \hline
    \end{tabular}
\caption{ \label{tab:distance-auslan} Auslan - Hamming Distance of $M$ ($95\times 95$)}
\end{center}
\end{table}

\begin{table}[htbp]
\begin{center}
    \begin{tabular}{|c|c|c|c|c|c|} 
    \hline
        \multicolumn{2}{|c|}{} & \multicolumn{2}{|c|}{$N$-ECOC Det} & \multicolumn{2}{|c|}{$N$-ECOC Rand} \\
    \hline 
        $N$ & $n_k$ & $d_r(M)$ & $d_T(M)$ & $d_r(M_R)$ & $d_T(M_R)$ \\
        \hline
        2 & 1024 & 496 & 992 & 435 & 864  \\
        3 & 2187 & 514 & 1028 & 601 & 1204  \\
        5 & 3125 & 625 & 1250 & 744 & 1487   \\
        7 & 2401 & 657 & 1314 & 808 & 1615   \\
        11 & 1331 & 879 & 1758 & 870 & 1736   \\
        13 & 2197 & 831 & 1662 & 885 & 1770  \\
    \hline
    \end{tabular}
\caption{\label{tab:distance-aloi} Aloi - ECOC matrix $M$ ($1000\times 1000$)}
\end{center}
\end{table}

\subsubsection{Comparing Minimum Distances of ECOC matrices for Deterministic vs Random $N$-ECOC (Absolute Distance)}

We provide additional results by reporting absolute row distances $d_r^{(a)}(M)$ and $d_r^{(a)}(M_A)$, and absolute total distances $d_r^{(a)}(M_A)$ and $d_T^{(a)}(M_A)$.  Here, $M_A$ refers to the ECOC matrix (referred to as a random-absolute $N$-ECOC matrix) chosen from the best of 1000 randomly generated matrices by optimzing absolute total distance, $d_T^{(a)}(M_A)$, where $d_T^{(a)}=d_r^{(a)}+d_c^{(a)}$ refers to total distance calculated using the absolute distance function defined by (7) in the main paper.  Results are reported for all datasets, except for Aloi because of the high computational cost.  

\

\noindent 1. Tables comparing absolute row distance and absolute total distance between deterministic versus random-absolute ECOC matrices for each dataset (except Aloi)

\begin{table}[htbp] 
\begin{center}
    \begin{tabular}{|c|c|c|c|c|c|} 
    \hline
        \multicolumn{2}{|c|}{} & \multicolumn{2}{|c|}{$N$-ECOC Det} &  \multicolumn{2}{|c|}{$N$-ECOC Rand-Abs} \\
    \hline 
        $N$ & $n_k$ & $d_r^{(a)}(M)$ &  $d_T^{(a)}(M)$ &  $d_r^{(a)}(M_A)$ & $d_T^{(a)}(M_A)$ \\
        \hline
        2 & 16 & 4 & 8 & 3 & 6 \\
        3 & 27 & 6 & 12 & 7 & 13 \\
        5 & 25 & 5 & 10 & 12 & 23 \\
        7 & 49 & 7 & 14 & 17 & 35 \\
    \hline
    \end{tabular}
\caption{\label{tab:distance-pendigits-abs} Pendigits - Absolute Distance of $M$ ($10\times 10$)}
\end{center}
\end{table}

\begin{table}[htbp] 
\begin{center}
    \begin{tabular}{|c|c|c|c|c|c|} 
    \hline
        \multicolumn{2}{|c|}{} & \multicolumn{2}{|c|}{$N$-ECOC Det} &  \multicolumn{2}{|c|}{$N$-ECOC Rand-Abs} \\
    \hline 
        $N$ & $n_k$ & $d_r^{(a)}(M)$ &  $d_T^{(a)}(M)$ &  $d_r^{(a)}(M_A)$ & $d_T^{(a)}(M_A)$ \\
        \hline
        2 & 16 & 4 & 8 & 3 & 6 \\
        3 & 27 & 6 & 12 & 6 & 12 \\
        5 & 25 & 5 & 10 & 11 & 22 \\
        7 & 49 & 7 & 14 & 17 & 34 \\
    \hline
    \end{tabular}
\caption{\label{tab:distance-usps-abs} Usps - Absolute Distance of $M$ ($10\times 10$)}
\end{center}
\end{table}

\begin{table}[htbp] 
\begin{center}
    \begin{tabular}{|c|c|c|c|c|c|} 
    \hline
        \multicolumn{2}{|c|}{} & \multicolumn{2}{|c|}{$N$-ECOC Det} &  \multicolumn{2}{|c|}{$N$-ECOC Rand-Abs} \\
    \hline 
        $N$ & $n_k$ & $d_r^{(a)}(M)$ &  $d_T^{(a)}(M)$ &  $d_r^{(a)}(M_A)$ & $d_T^{(a)}(M_A)$ \\
        \hline
        2 & 16 & 4 & 8 & 4 & 7 \\
        3 & 27 & 7 & 14 & 7 & 14 \\
        5 & 25 & 6 & 12 & 12 & 26\\
        7 & 49 & 7 & 14 & 18 & 36 \\
        11 & 11 & 31 & 62 & 29 & 57 \\
    \hline
    \end{tabular}
\caption{\label{tab:distance-vowel-abs} Vowel - Absolute Distance of $M$ ($11\times 11$)}
\end{center}
\end{table}

\begin{table}[htbp]
\begin{center}
    \begin{tabular}{|c|c|c|c|c|c|} 
    \hline
        \multicolumn{2}{|c|}{} & \multicolumn{2}{|c|}{$N$-ECOC Det} &  \multicolumn{2}{|c|}{$N$-ECOC Rand-Abs} \\
    \hline 
        $N$ & $n_k$ & $d_r^{(a)}(M)$ &  $d_T^{(a)}(M)$ &  $d_r^{(a)}(M_A)$ & $d_T^{(a)}(M_A)$ \\
        \hline
        2 & 32 & 12 & 24 & 8 & 16\\
        3 & 27 & 20 & 40 & 16 & 31 \\
        5 & 125 & 30 & 60 & 29 & 59 \\
        7 & 49 & 30 & 60 & 43 & 84  \\
        11 & 121 & 15 & 30 & 68 & 134\\
        13 & 169 & 13 & 26 & 85 & 161 \\
    \hline
    \end{tabular}
\caption{\label{tab:distance-letters-abs} Letters - Absolute Distance of $M$ ($26\times 26$)}
\end{center}
\end{table}

\begin{table}[htbp]
\begin{center}
    \begin{tabular}{|c|c|c|c|c|c|} 
    \hline
        \multicolumn{2}{|c|}{} & \multicolumn{2}{|c|}{$N$-ECOC Det} &  \multicolumn{2}{|c|}{$N$-ECOC Rand-Abs} \\
    \hline 
        $N$ & $n_k$ & $d_r^{(a)}(M)$ &  $d_T^{(a)}(M)$ &  $d_r^{(a)}(M_A)$ & $d_T^{(a)}(M_A)$ \\
        \hline
        2 & 128 & 46 & 92 & 33 & 66 \\
        3 & 243 & 69 & 138 & 62 & 126 \\
        5 & 125 & 95 & 95 & 116 & 236 \\
        7 & 343 & 54 & 108 & 172 & 343 \\
        11 & 121 & 95 & 526 & 279 & 552 \\
        13 & 169 & 190 & 470 & 323 & 649 \\
    \hline
    \end{tabular}
\caption{ \label{tab:distance-auslan-abs} Auslan - Absolute Distance of $M$ ($95\times 95$)}
\end{center}
\end{table}

\subsubsection{Accuracy between Deterministic vs Random $N$-ECOC (Hamming Distance)}

We first present accuracy results comparing three strategies: deterministic $N$-ECOC, random $N$-ECOC, and random-row $N$-ECOC.  For random and random-row strategies, ECOC matrices were chosen from best of 1000 by optimizing total distance and row distance, respectively, using the Hamming distance function defined by (6) in the main paper.

\

\noindent  1. Comparison of accuracy between deterministic $N$-ECOC, random $N$-ECOC, and random-row $N$-ECOC using DT for square matrix dimension (Figures 1-6):

\begin{figure}[htbp]
\centerline{\includegraphics[width=140pt,keepaspectratio]{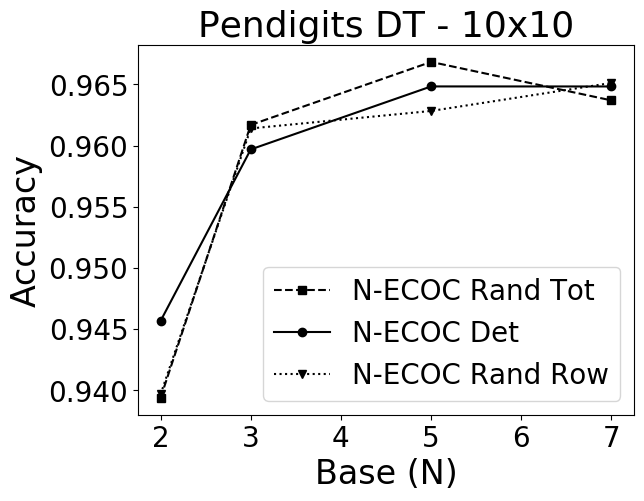}}
\caption{Deterministic and Random $N$-ECOC for Pendigits using DT: Base vs. Accuracy (square)}
\label{fig:pendigits-det-vs-rand-DT}
\end{figure}

\begin{figure}[htbp]
\centerline{\includegraphics[width=140pt,keepaspectratio]{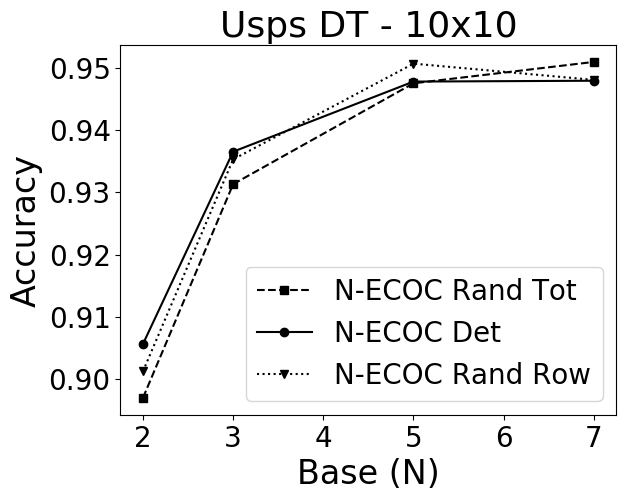}}
\caption{Deterministic and Random $N$-ECOC for Usps using DT: Base vs. Accuracy (square)}
\label{fig:usps-det-vs-rand-DT}
\end{figure}

\begin{figure}[htbp]
\centerline{\includegraphics[width=140pt,keepaspectratio]{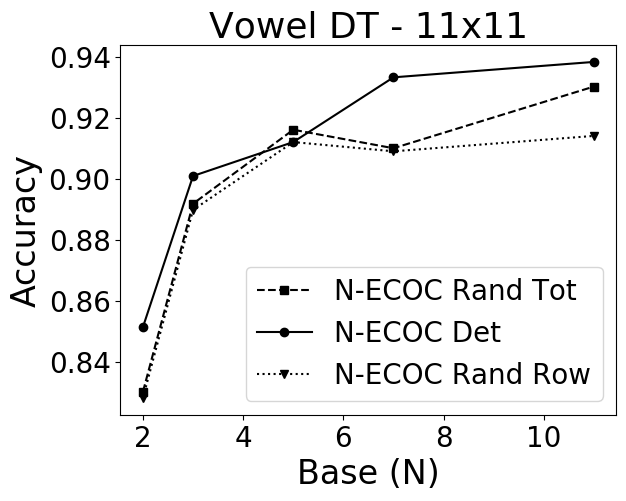}}
\caption{Deterministic and Random $N$-ECOC for Vowel using DT: Base vs. Accuracy (square)}
\label{fig:vowel-det-vs-rand-DT}
\end{figure}

\begin{figure}[htbp]
\centerline{\includegraphics[width=140pt,keepaspectratio]{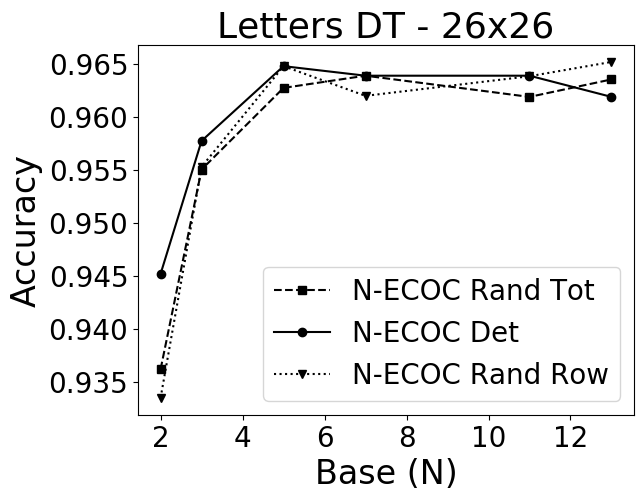}}
\caption{Deterministic and Random $N$-ECOC for Letters using DT: Base vs. Accuracy (square)}
\label{fig:letter-det-vs-rand-DT}
\end{figure}

\begin{figure}[htbp]
\centerline{\includegraphics[width=140pt,keepaspectratio]{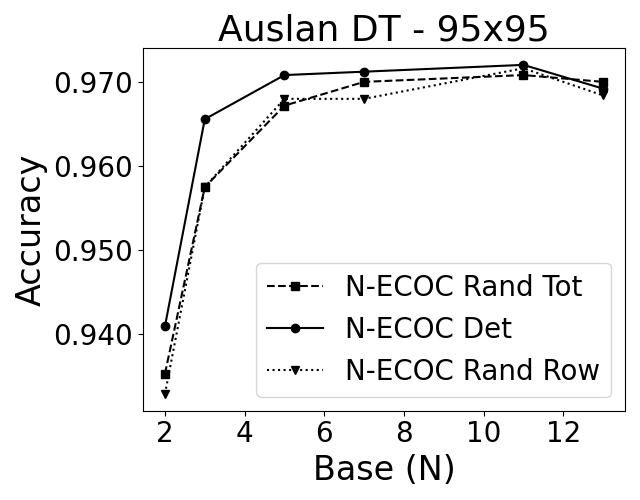}}
\caption{Deterministic and Random $N$-ECOC for Auslan using DT: Base vs. Accuracy (square)}
\label{fig:auslan-det-vs-rand-DT}
\end{figure}

\begin{figure}[htbp]
\centerline{\includegraphics[width=140pt,keepaspectratio]{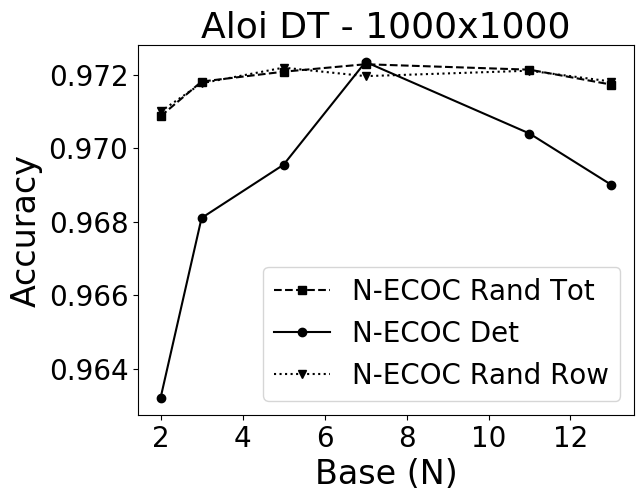}}
\caption{Deterministic and Random $N$-ECOC for Aloi using DT: Base vs. Accuracy (square)}
\label{fig:aloi-det-vs-rand-DT}
\end{figure}

\noindent  2. Comparison of accuracy between deterministic $N$-ECOC, random $N$-ECOC, and random-row $N$-ECOC using SVM for square matrix dimension (Figures 7-12)

\begin{figure}[htbp]
\centerline{\includegraphics[width=140pt,keepaspectratio]{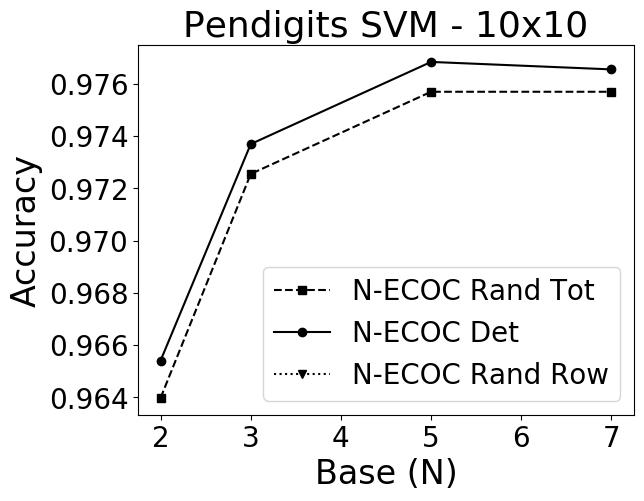}}
\caption{Deterministic and Random $N$-ECOC for Pendigits using SVM: Base vs. Accuracy (square)}
\label{fig:pendigits-det-vs-rand-SVM}
\end{figure}

\begin{figure}[htbp]
\centerline{\includegraphics[width=140pt,keepaspectratio]{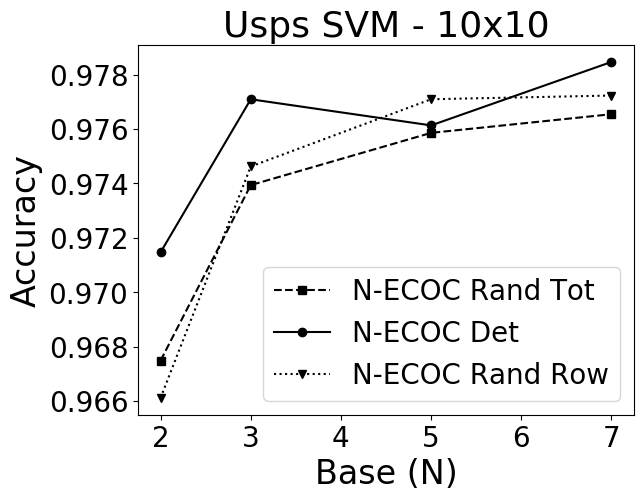}}
\caption{Deterministic and Random $N$-ECOC for Usps using SVM: Base vs. Accuracy (square)}
\label{fig:usps-det-vs-rand-SVM}
\end{figure}

\begin{figure}[htbp]
\centerline{\includegraphics[width=140pt,keepaspectratio]{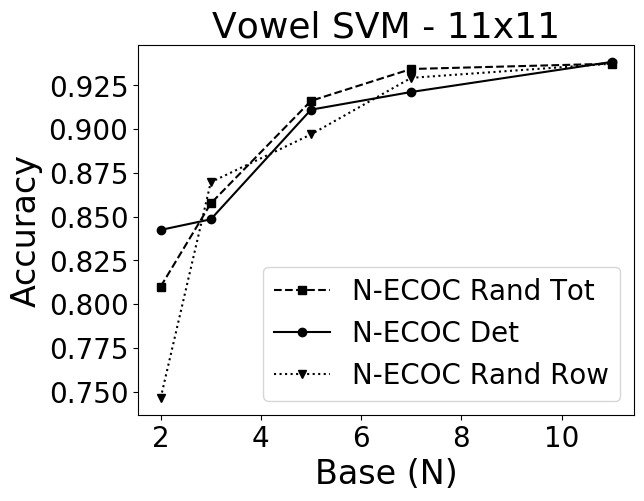}}
\caption{Deterministic and Random $N$-ECOC for Vowel using SVM: Base vs. Accuracy (square)}
\label{fig:vowel-det-vs-rand-SVM}
\end{figure}

\begin{figure}[htbp]
\centerline{\includegraphics[width=140pt,keepaspectratio]{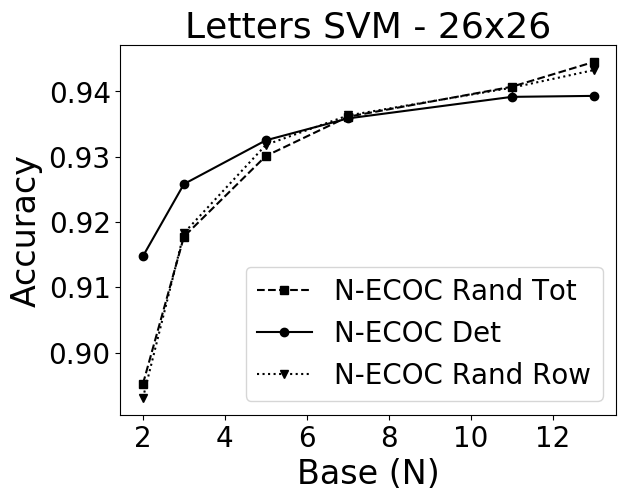}}
\caption{Deterministic and Random $N$-ECOC for Letters using SVM: Base vs. Accuracy (square)}
\label{fig:letter-det-vs-rand-SVM}
\end{figure}

\begin{figure}[htbp]
\centerline{\includegraphics[width=140pt,keepaspectratio]{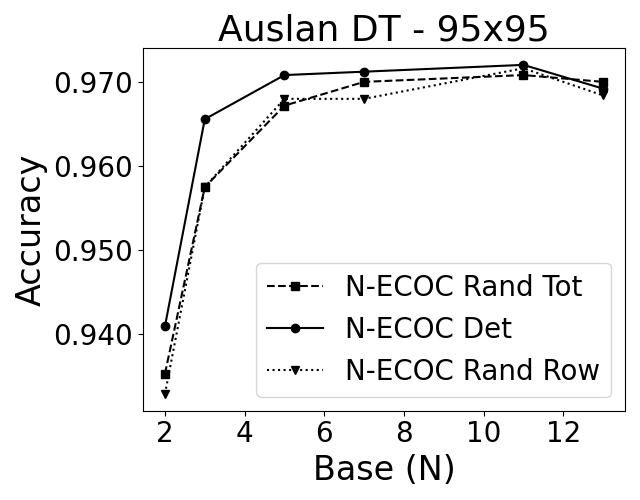}}
\caption{Deterministic and Random $N$-ECOC for Auslan using SVM: Base vs. Accuracy (square)}
\label{fig:auslan-det-vs-rand-SVM}
\end{figure}

\begin{figure}[htbp]
\centerline{\includegraphics[width=140pt,keepaspectratio]{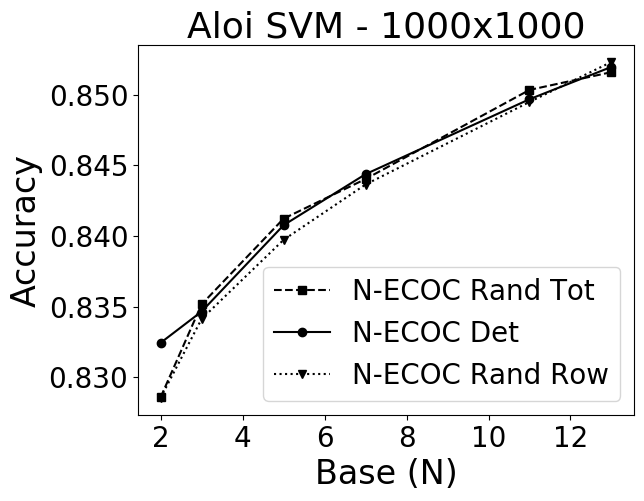}}
\caption{Deterministic and Random $N$-ECOC for Aloi using SVM: Base vs. Accuracy (square)}
\label{fig:aloi-det-vs-rand-SVM}
\end{figure}


We next present similar results but for half and double matrix dimensions.

\

\noindent  3. Comparison of accuracy between deterministic $N$-ECOC, random $N$-ECOC, and rando-row $N$-ECOC using DT for half matrix dimension (Figures 13-18)

\begin{figure}[htbp]
\centerline{\includegraphics[width=140pt,keepaspectratio]{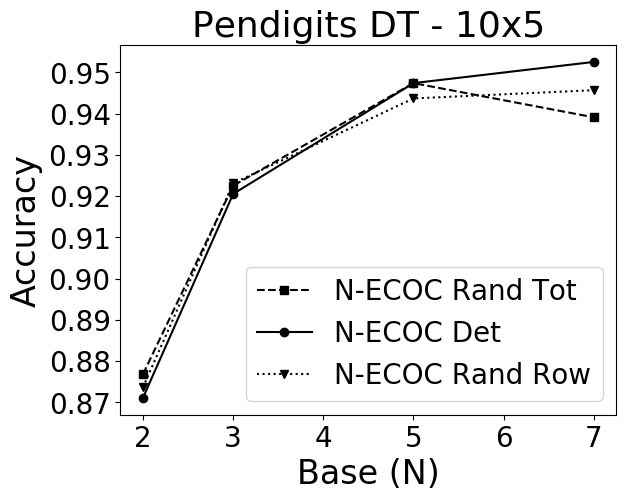}}
\caption{Deterministic and Random $N$-ECOC for Pendigits using DT: Base vs. Accuracy (half)}
\label{fig:pendigits-det-vs-rand-half-DT}
\end{figure}

\begin{figure}[htbp]
\centerline{\includegraphics[width=140pt,keepaspectratio]{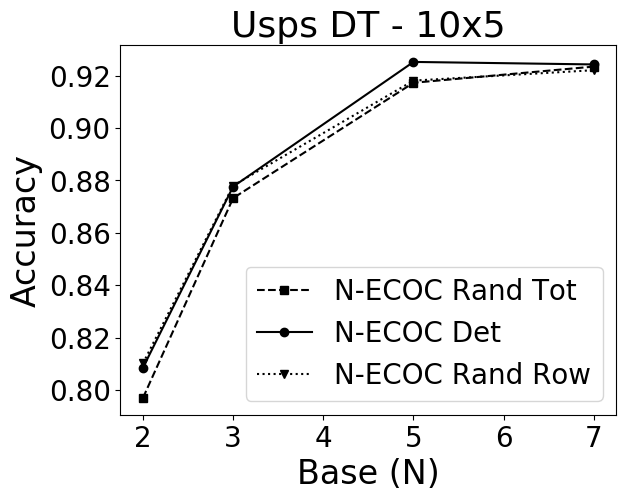}}
\caption{Deterministic and Random $N$-ECOC for Usps using DT: Base vs. Accuracy (half)}
\label{fig:usps-det-vs-rand-half-DT}
\end{figure}

\begin{figure}[htbp]
\centerline{\includegraphics[width=140pt,keepaspectratio]{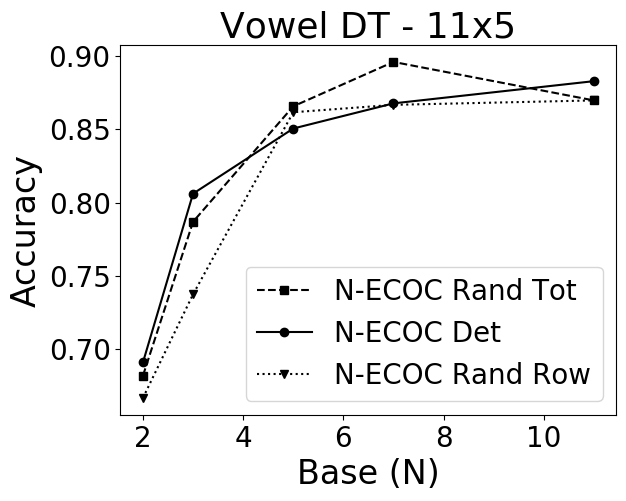}}
\caption{Deterministic and Random $N$-ECOC for Vowel using DT: Base vs. Accuracy (half)}
\label{fig:vowel-det-vs-rand-half-DT}
\end{figure}

\begin{figure}[htbp]
\centerline{\includegraphics[width=140pt,keepaspectratio]{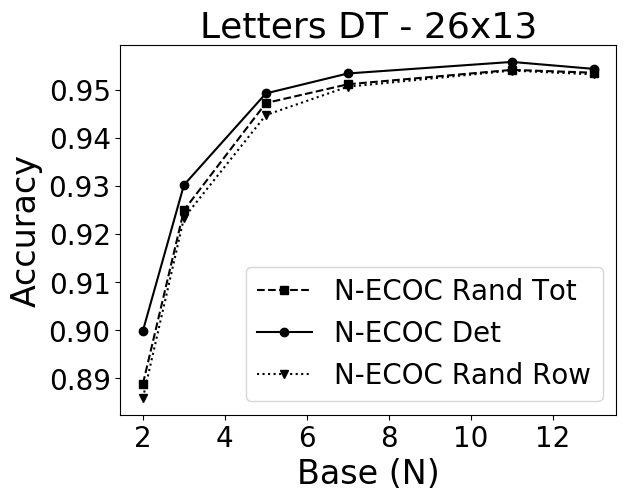}}
\caption{Deterministic and Random $N$-ECOC for Letters using DT: Base vs. Accuracy (half)}
\label{fig:letter-det-vs-rand-half-DT}
\end{figure}

\begin{figure}[htbp]
\centerline{\includegraphics[width=140pt,keepaspectratio]{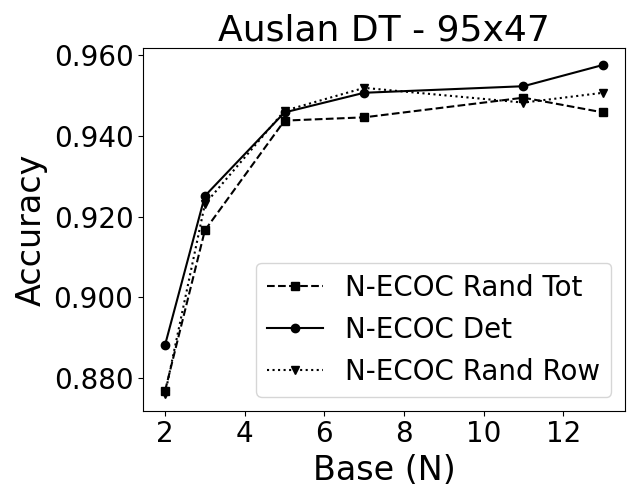}}
\caption{Deterministic and Random $N$-ECOC for Auslan using DT: Base vs. Accuracy (half)}
\label{fig:auslan-det-vs-rand-half-DT}
\end{figure}

\begin{figure}[htbp]
\centerline{\includegraphics[width=140pt,keepaspectratio]{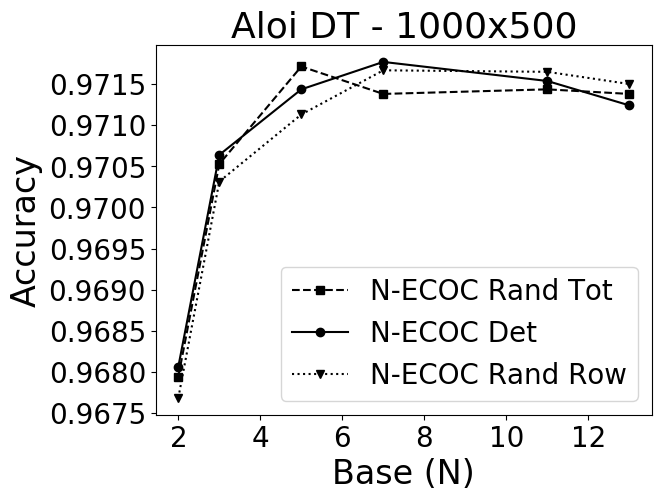}}
\caption{Deterministic and Random $N$-ECOC for Aloi using DT: Base vs. Accuracy (half)}
\label{fig:aloi-det-vs-rand-half-DT}
\end{figure}

\noindent  4. Comparison of accuracy between deterministic $N$-ECOC, random $N$-ECOC, and rando-row $N$-ECOC using SVM for half matrix dimension (Figures 19-24)

\begin{figure}[htbp]
\centerline{\includegraphics[width=140pt,keepaspectratio]{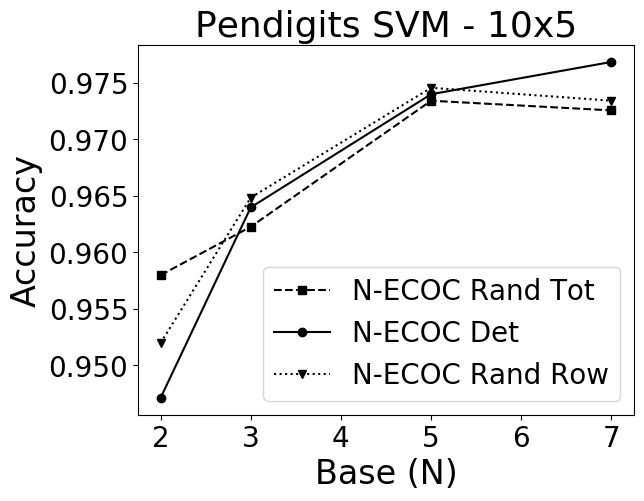}}
\caption{Deterministic and Random $N$-ECOC for Pendigits using SVM: Base vs. Accuracy (half)}
\label{fig:pendigits-det-vs-rand-half-SVM}
\end{figure}

\begin{figure}[htbp]
\centerline{\includegraphics[width=140pt,keepaspectratio]{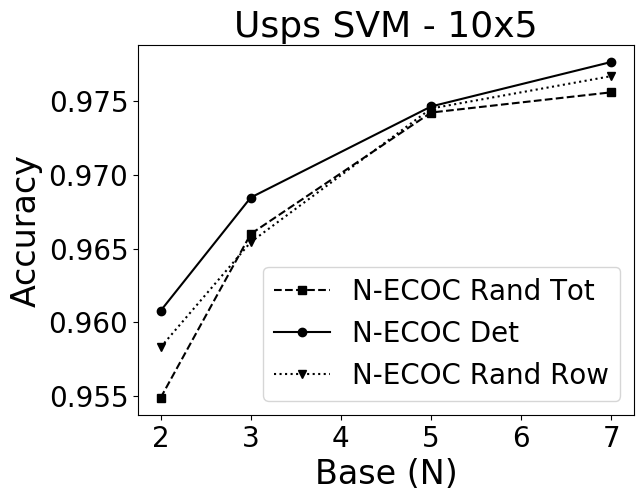}}
\caption{Deterministic and Random $N$-ECOC for Usps using SVM: Base vs. Accuracy (half)}
\label{fig:usps-det-vs-rand-half-SVM}
\end{figure}

\begin{figure}[htbp]
\centerline{\includegraphics[width=140pt,keepaspectratio]{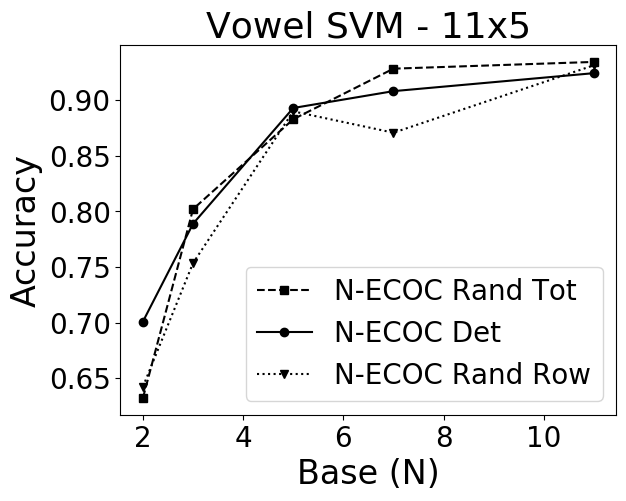}}
\caption{Deterministic and Random $N$-ECOC for Vowel using SVM: Base vs. Accuracy (half)}
\label{fig:vowel-det-vs-rand-half-SVM}
\end{figure}

\begin{figure}[htbp]
\centerline{\includegraphics[width=140pt,keepaspectratio]{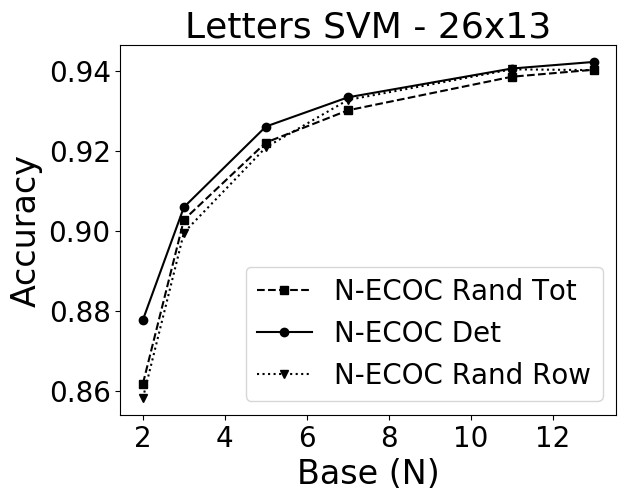}}
\caption{Deterministic and Random $N$-ECOC for Letters using SVM: Base vs. Accuracy (half)}
\label{fig:letter-det-vs-rand-half-SVM}
\end{figure}

\begin{figure}[htbp]
\centerline{\includegraphics[width=140pt,keepaspectratio]{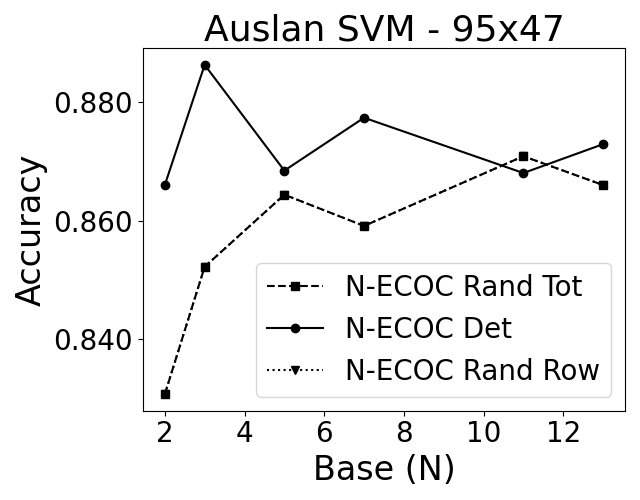}}
\caption{Deterministic and Random $N$-ECOC for Auslan using SVM: Base vs. Accuracy (half)}
\label{fig:auslan-det-vs-rand-half-SVM}
\end{figure}

\begin{figure}[htbp]
\centerline{\includegraphics[width=140pt,keepaspectratio]{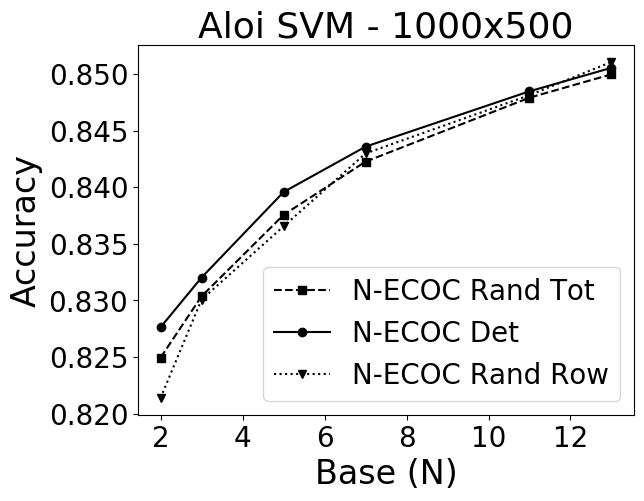}}
\caption{Deterministic and Random $N$-ECOC for Aloi using SVM: Base vs. Accuracy (half)}
\label{fig:aloi-det-vs-rand-half-SVM}
\end{figure}

\noindent  5. Comparison of accuracy between deterministic $N$-ECOC, random $N$-ECOC, and rando-row $N$-ECOC using DT for double matrix dimension (Figures 25-29).  No results for Aloi due to the expensive computational cost.

\begin{figure}[htbp]
\centerline{\includegraphics[width=140pt,keepaspectratio]{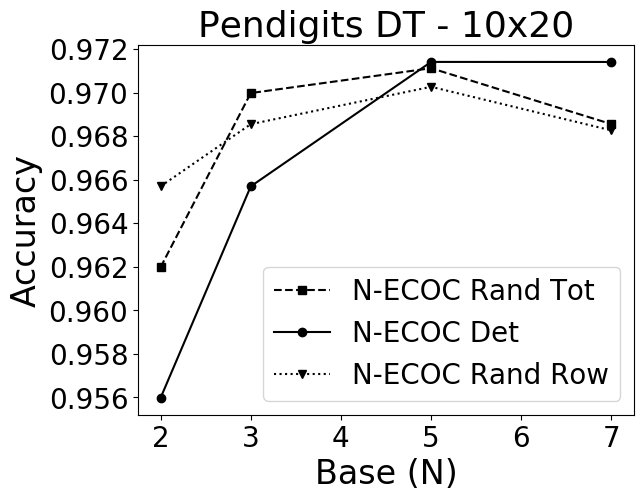}}
\caption{Deterministic and Random $N$-ECOC for Pendigits using DT: Base vs. Accuracy (double)}
\label{fig:pendigits-det-vs-rand-double-DT}
\end{figure}

\begin{figure}[htbp]
\centerline{\includegraphics[width=140pt,keepaspectratio]{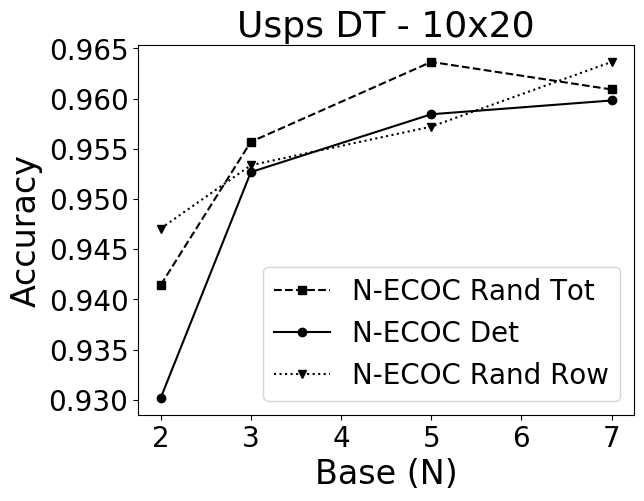}}
\caption{Deterministic and Random $N$-ECOC for Usps using DT: Base vs. Accuracy (double)}
\label{fig:usps-det-vs-rand-double-DT}
\end{figure}

\begin{figure}[htbp]
\centerline{\includegraphics[width=140pt,keepaspectratio]{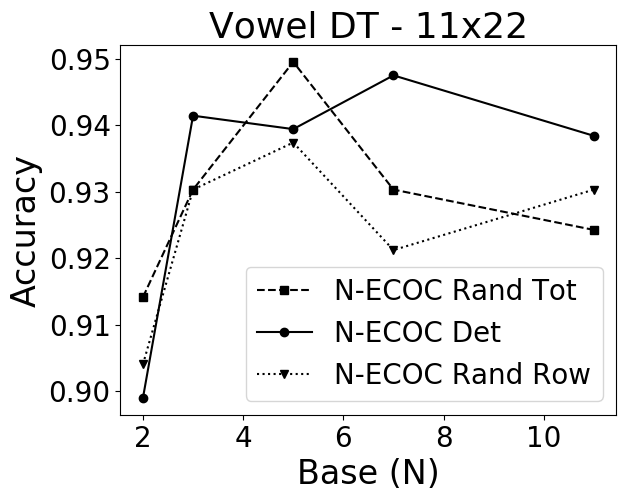}}
\caption{Deterministic and Random $N$-ECOC for Vowel using DT: Base vs. Accuracy (double)}
\label{fig:vowel-det-vs-rand-double-DT}
\end{figure}

\begin{figure}[htbp]
\centerline{\includegraphics[width=140pt,keepaspectratio]{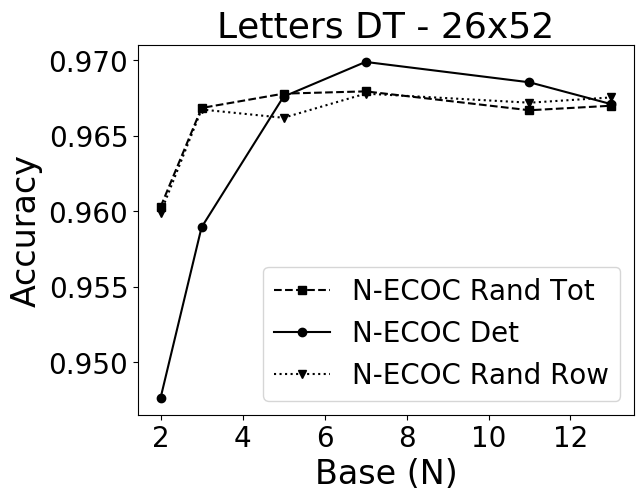}}
\caption{Deterministic and Random $N$-ECOC for Letters using DT: Base vs. Accuracy (double)}
\label{fig:letter-det-vs-rand-double-DT}
\end{figure}

\begin{figure}[htbp]
\centerline{\includegraphics[width=140pt,keepaspectratio]{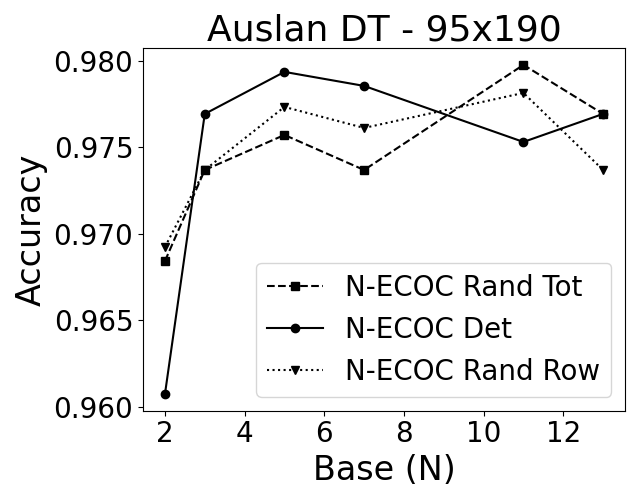}}
\caption{Deterministic and Random $N$-ECOC for Auslan using DT: Base vs. Accuracy (double)}
\label{fig:auslan-det-vs-rand-double-DT}
\end{figure}

6. Comparison of accuracy between deterministic $N$-ECOC, random $N$-ECOC, and rando-row $N$-ECOC using SVM for double matrix dimension (Figures 30-34).  No results for Aloi due to the expensive computational cost.

\begin{figure}[htbp]
\centerline{\includegraphics[width=140pt,keepaspectratio]{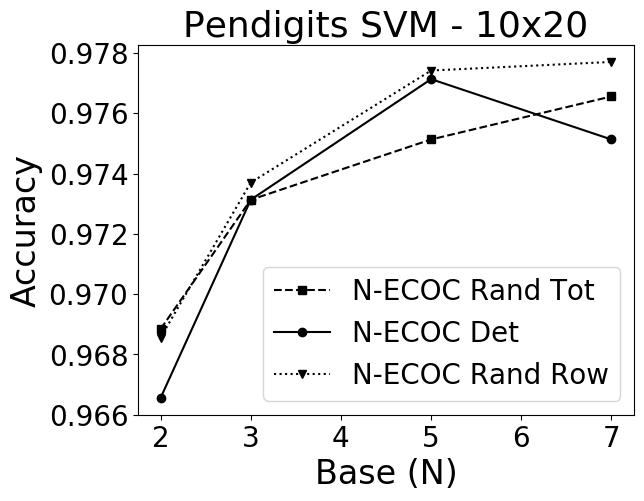}}
\caption{Deterministic and Random $N$-ECOC for Pendigits using SVM: Base vs. Accuracy (double)}
\label{fig:pendigits-det-vs-rand-double-SVM}
\end{figure}

\begin{figure}[htbp]
\centerline{\includegraphics[width=140pt,keepaspectratio]{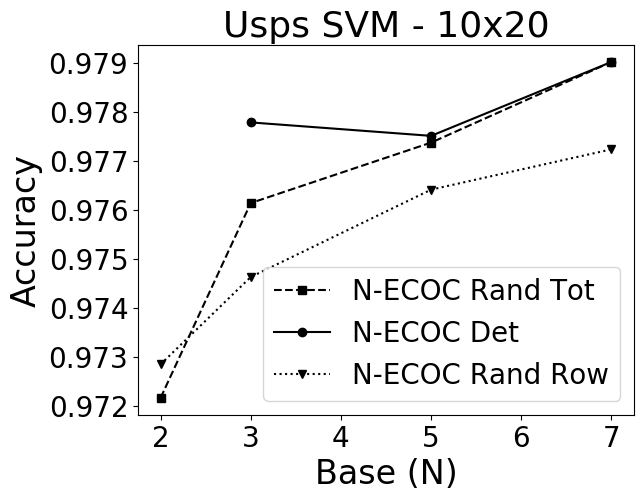}}
\caption{Deterministic and Random $N$-ECOC for Usps using SVM: Base vs. Accuracy (double)}
\label{fig:usps-det-vs-rand-double-SVM}
\end{figure}

\begin{figure}[htbp]
\centerline{\includegraphics[width=140pt,keepaspectratio]{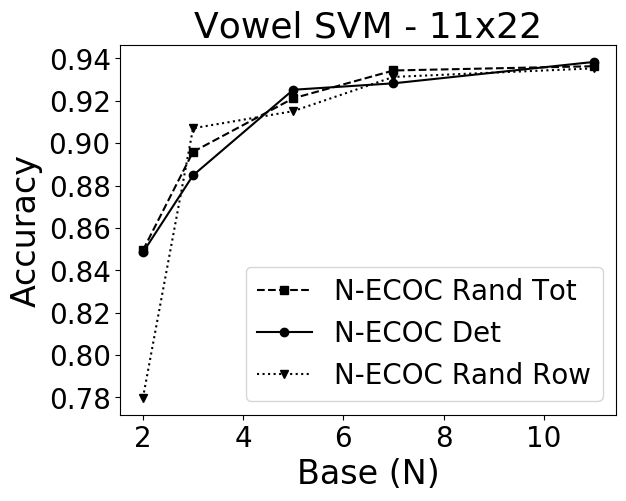}}
\caption{Deterministic and Random $N$-ECOC for Vowel using SVM: Base vs. Accuracy (double)}
\label{fig:vowel-det-vs-rand-double-SVM}
\end{figure}

\begin{figure}[htbp]
\centerline{\includegraphics[width=140pt,keepaspectratio]{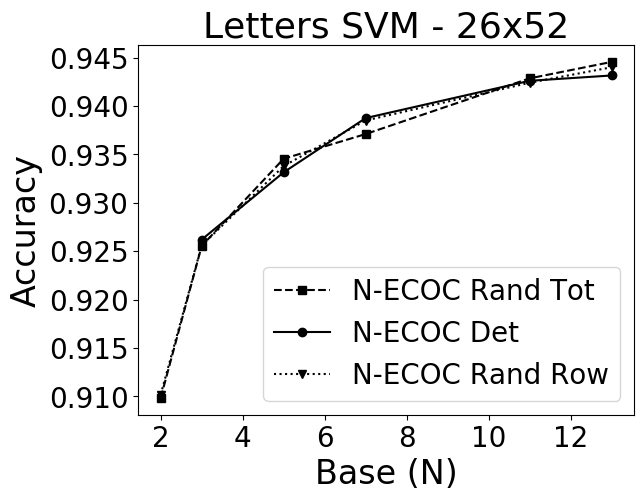}}
\caption{Deterministic and Random $N$-ECOC for Letters using SVM: Base vs. Accuracy (double)}
\label{fig:letter-det-vs-rand-double-SVM}
\end{figure}

\begin{figure}[htbp]
\centerline{\includegraphics[width=140pt,keepaspectratio]{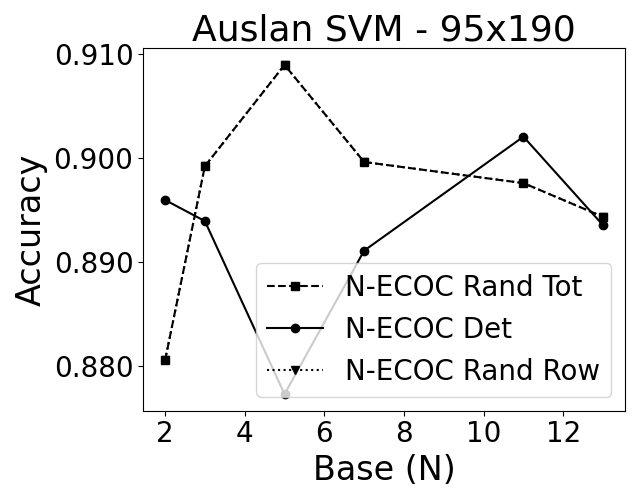}}
\caption{Deterministic and Random $N$-ECOC for Auslan using SVM: Base vs. Accuracy (double)}
\label{fig:auslan-det-vs-rand-double-SVM}
\end{figure}

\subsubsection{Accuracy between Deterministic vs Random $N$-ECOC (Hamming Distance and Absolute Distance)}

In this section we report accuracy results for three strategies:  deterministic $N$-ECOC, random $N$-ECOC (Hamming ddistance), and random-absolute $N$-ECOC (absolute distance).  For the random (Ham $N$-ECOC Rand) and random-absolute (Abs $N$-ECOC Rand) strategies, ECOC matrices $M_R$ and $M_A$ were chosen from best of 1000 randomly generated matrices by optimizing $d_T^(M_R)$ and $d_T^{(a)}(M_A)$ based on the distance functions (6) and (7), respectively, in the main paper.  We only report results for square ECOC matrices and only for the datasets Pendigits, Usps, and Letters, which we believe sufficiently shows that there is no clear significant difference in accuracy between the random and random-absolute strategies.

\

\noindent  1. Comparison of accuracy between deterministic, random, and random-absolute $N$-ECOC using DT for square matrix dimension (Figures \ref{fig:pendigits-det-vs-rand-abs-DT}-\ref{fig:letter-det-vs-rand-abs-DT}):

\begin{figure}[htbp]
\centerline{\includegraphics[width=140pt,keepaspectratio]{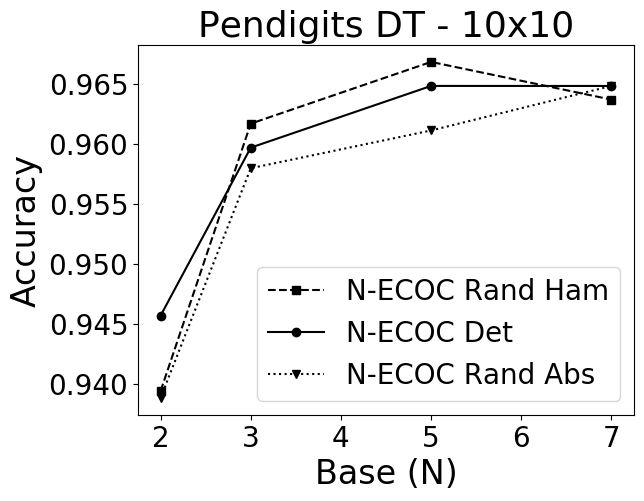}}
\caption{Deterministic and Random (Hamming and Absolute) $N$-ECOC for Pendigits using DT: Base vs. Accuracy (square)}
\label{fig:pendigits-det-vs-rand-abs-DT}
\end{figure}

\begin{figure}[htbp]
\centerline{\includegraphics[width=140pt,keepaspectratio]{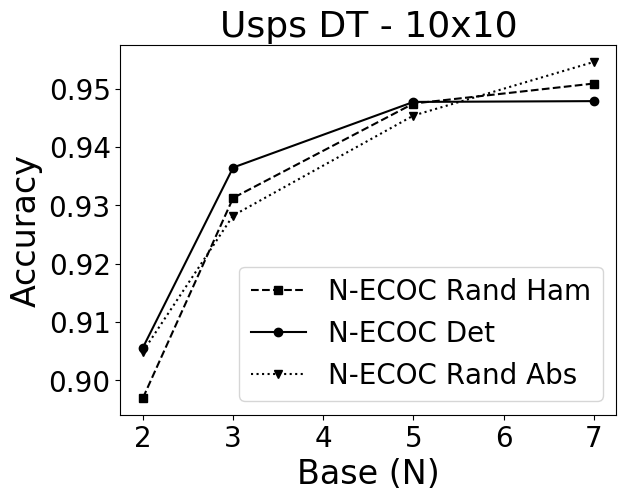}}
\caption{Deterministic and Random (Hamming and Absolute) $N$-ECOC for Usps using DT: Base vs. Accuracy (square)}
\label{fig:usps-det-vs-rand-abs-DT}
\end{figure}

\begin{figure}[htbp]
\centerline{\includegraphics[width=140pt,keepaspectratio]{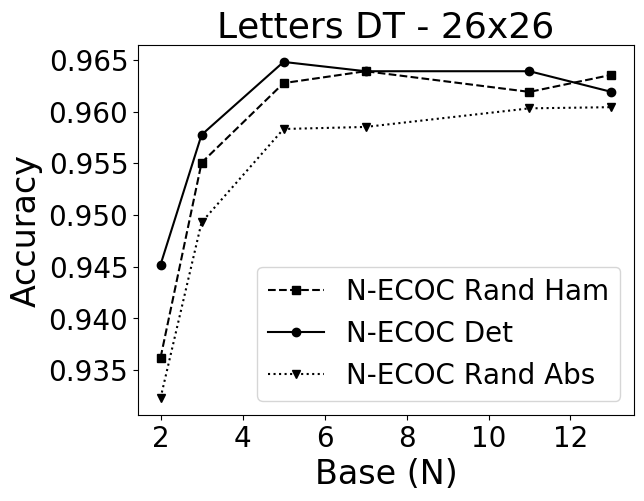}}
\caption{Deterministic and Random (Hamming and Absolute) $N$-ECOC for Letters using DT: Base vs. Accuracy (square)}
\label{fig:letter-det-vs-rand-abs-DT}
\end{figure}

\noindent  2. Comparison of accuracy between deterministic, random, and random-absolute $N$-ECOC using SVM for square matrix dimension (Figures \ref{fig:pendigits-det-vs-rand-abs-SVM}-\ref{fig:letter-det-vs-rand-abs-SVM}):  

\begin{figure}[htbp]
\centerline{\includegraphics[width=140pt,keepaspectratio]{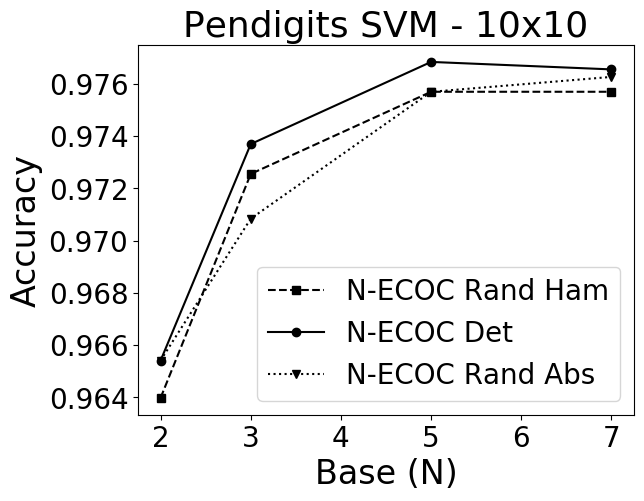}}
\caption{Deterministic and Random (Hamming and Absolute) $N$-ECOC for Pendigits using SVM: Base vs. Accuracy (square)}
\label{fig:pendigits-det-vs-rand-abs-SVM}
\end{figure}

\begin{figure}[htbp]
\centerline{\includegraphics[width=140pt,keepaspectratio]{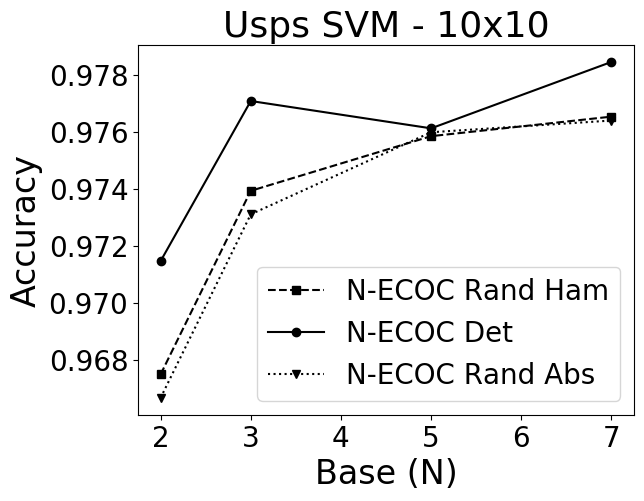}}
\caption{Deterministic and Random (Hamming and Absolute) $N$-ECOC for Usps using SVM: Base vs. Accuracy (square)}
\label{fig:usps-det-vs-rand-abs-SVM}
\end{figure}

\begin{figure}[htbp]
\centerline{\includegraphics[width=140pt,keepaspectratio]{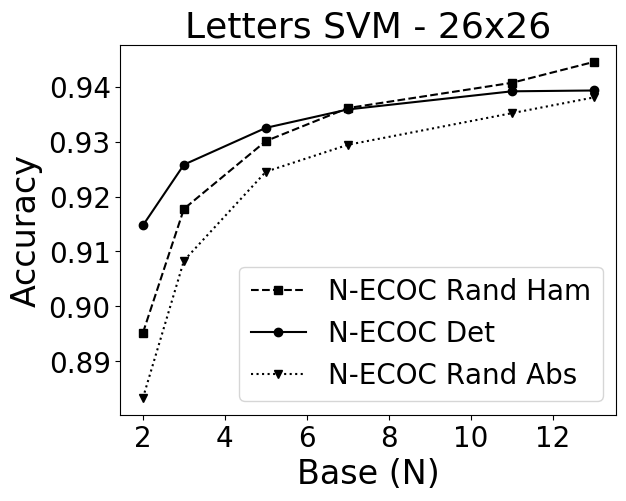}}
\caption{Deterministic and Random (Hamming and Absolute) $N$-ECOC for Letters using SVM: Base vs. Accuracy (square)}
\label{fig:letter-det-vs-rand-abs-SVM}
\end{figure}


\subsubsection{Varying Codeword Length}

In this section we present results comparing different matrix dimensions (half, square, and double) for the deterministic and random $N$-ECOC strategies.  For Aloi, we only report results for square and half dimensions due to the high computational cost.

\

\begin{figure}
\centerline{\includegraphics[width=140pt,keepaspectratio]{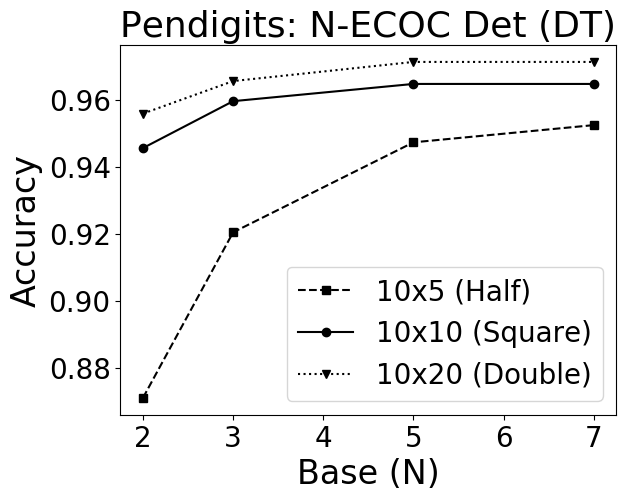}}
\caption{Deterministic $N$-ECOC for Pendigits using Half, Square, and Double dimensions: Base vs. Accuracy (DT)}
\label{fig:pendigits-length-DT}
\end{figure}

\begin{figure}
\centerline{\includegraphics[width=140pt,keepaspectratio]{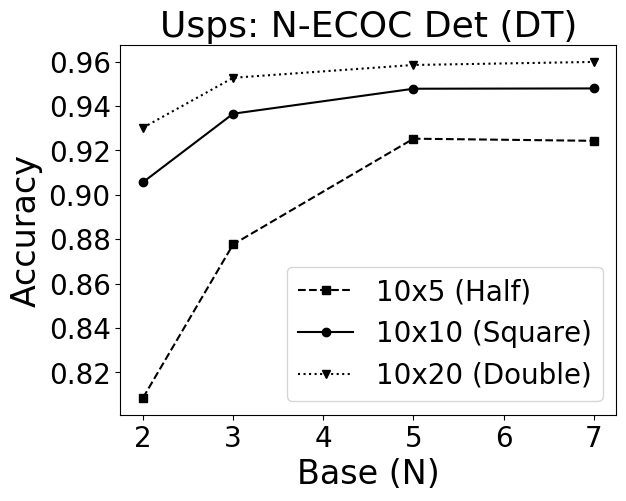}}
\caption{Deterministic $N$-ECOC for Usps using Half, Square, and Double dimensions: Base vs. Accuracy (DT)}
\label{fig:usps-length-DT}
\end{figure}

\begin{figure}
\centerline{\includegraphics[width=140pt,keepaspectratio]{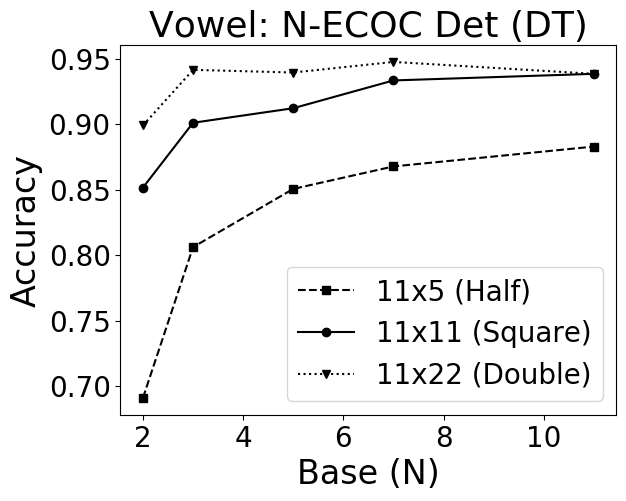}}
\caption{Deterministic $N$-ECOC for Vowel using Half, Square, and Double dimensions: Base vs. Accuracy (DT)}
\label{fig:vowel-length-DT}
\end{figure}

\begin{figure}
\centerline{\includegraphics[width=140pt,keepaspectratio]{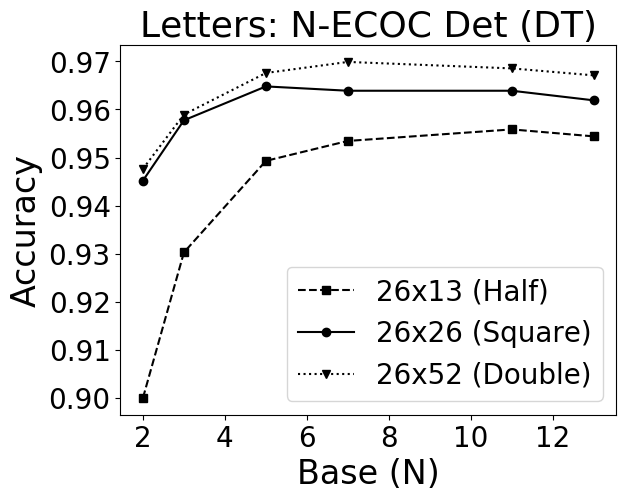}}
\caption{Deterministic $N$-ECOC for Letters using Half, Square, and Double dimensions: Base vs. Accuracy (DT)}
\label{fig:letters-length-DT}
\end{figure}

\begin{figure}
\centerline{\includegraphics[width=140pt,keepaspectratio]{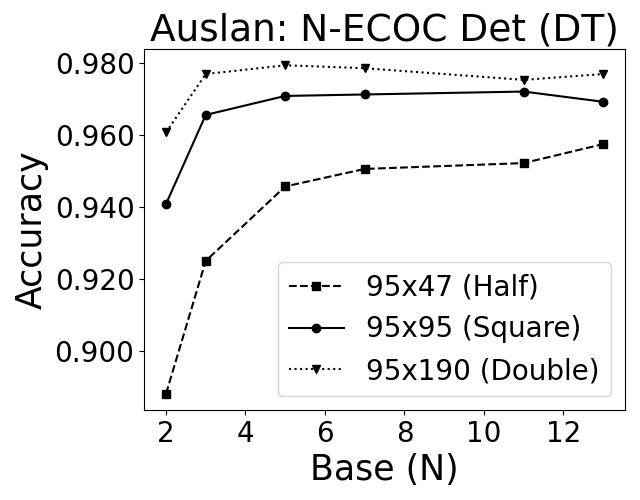}}
\caption{Deterministic $N$-ECOC for Auslan using Half, Square, and Double dimensions: Base vs. Accuracy (DT)}
\label{fig:auslan-length-DT}
\end{figure}

\begin{figure}
\centerline{\includegraphics[width=140pt,keepaspectratio]{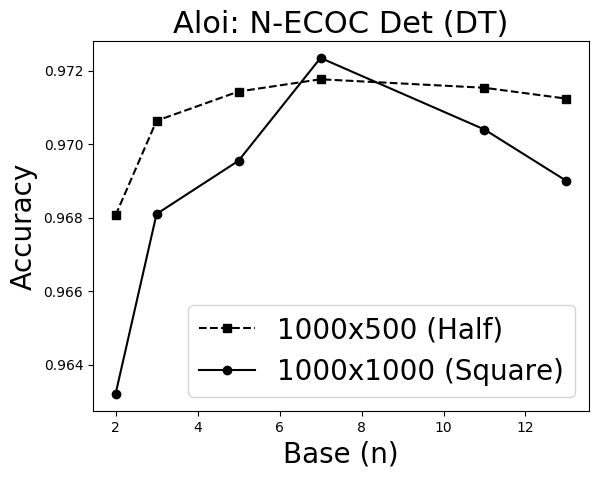}}
\caption{Deterministic $N$-ECOC for Aloi using Half, Square, and Double dimensions: Base vs. Accuracy (DT)}
\label{fig:aloi-length-DT}
\end{figure}

\noindent  1. Comparison of accuracy between half, square, and double for deterministic $N$-ECOC using DT (Figures \ref{fig:pendigits-length-DT}-\ref{fig:aloi-length-DT})

\begin{figure}
\centerline{\includegraphics[width=140pt,keepaspectratio]{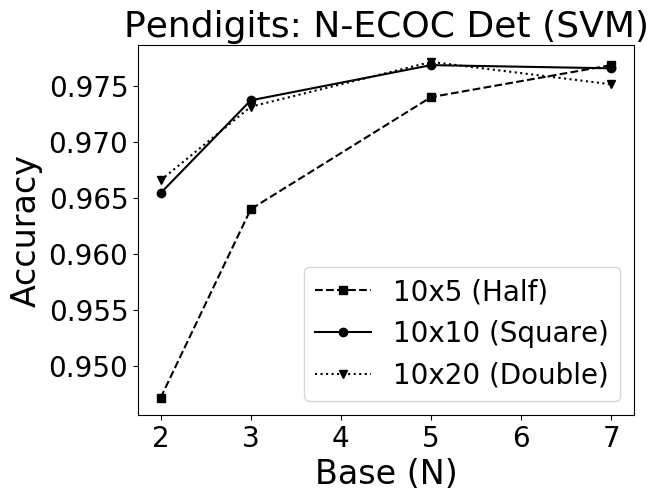}}
\caption{Deterministic $N$-ECOC for Pendigits using Half, Square, and Double dimensions: Base vs. Accuracy (SVM)}
\label{fig:pendigits-length-SVM}
\end{figure}

\begin{figure}
\centerline{\includegraphics[width=140pt,keepaspectratio]{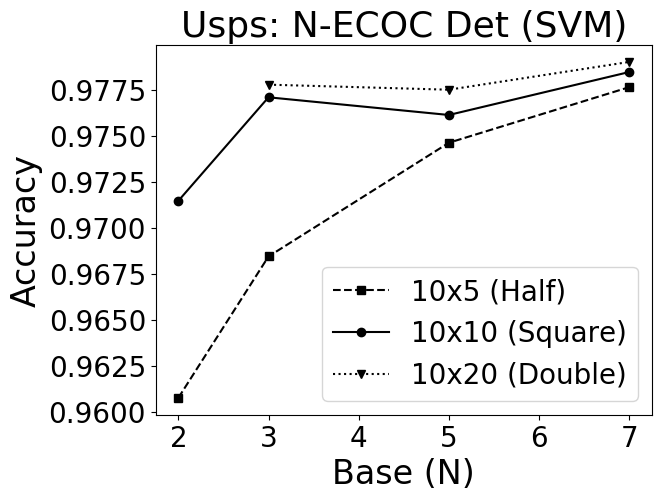}}
\caption{Deterministic $N$-ECOC for Usps using Half, Square, and Double dimensions: Base vs. Accuracy (SVM)}
\label{fig:usps-length-SVM}
\end{figure}

\begin{figure}
\centerline{\includegraphics[width=140pt,keepaspectratio]{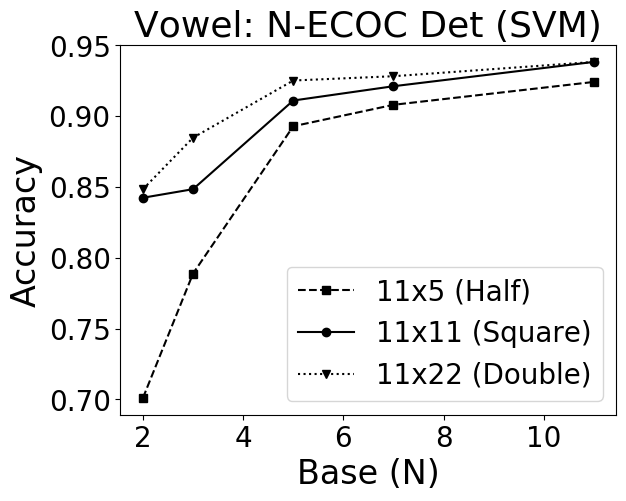}}
\caption{Deterministic $N$-ECOC for Vowel using Half, Square, and Double dimensions: Base vs. Accuracy (SVM)}
\label{fig:vowel-length-SVM}
\end{figure}

\begin{figure}
\centerline{\includegraphics[width=140pt,keepaspectratio]{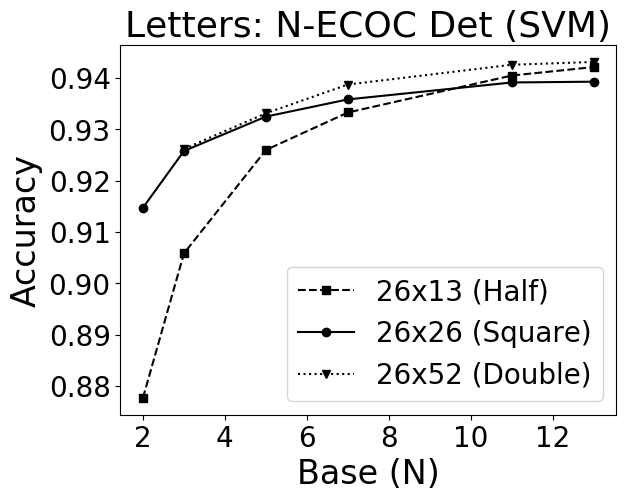}}
\caption{Deterministic $N$-ECOC for Letters using Half, Square, and Double dimensions: Base vs. Accuracy (SVM)}
\label{fig:letters-length-SVM}
\end{figure}

\begin{figure}
\centerline{\includegraphics[width=140pt,keepaspectratio]{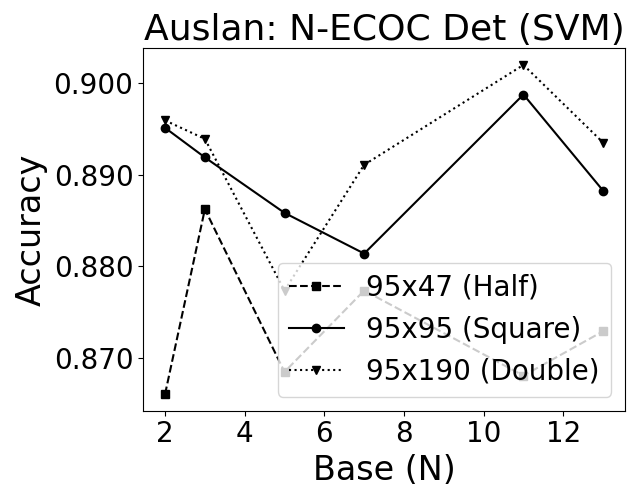}}
\caption{Deterministic $N$-ECOC for Auslan using Half, Square, and Double dimensions: Base vs. Accuracy (SVM)}
\label{fig:auslan-length-SVM}
\end{figure}

\begin{figure}
\centerline{\includegraphics[width=140pt,keepaspectratio]{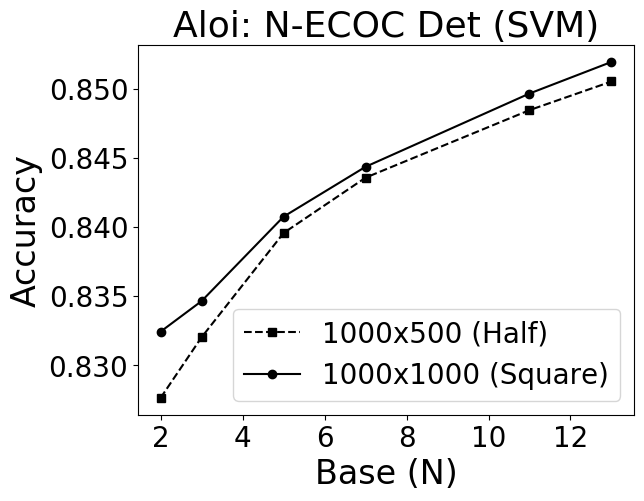}}
\caption{Deterministic $N$-ECOC for Aloi using Half, Square, and Double dimensions: Base vs. Accuracy (SVM)}
\label{fig:aloi-length-SVM}
\end{figure}

\noindent  2. Comparison of accuracy between half, square, and double for deterministic $N$-ECOC using SVM (Figures \ref{fig:pendigits-length-SVM}-\ref{fig:aloi-length-SVM})

\begin{figure}
\centerline{\includegraphics[width=140pt,keepaspectratio]{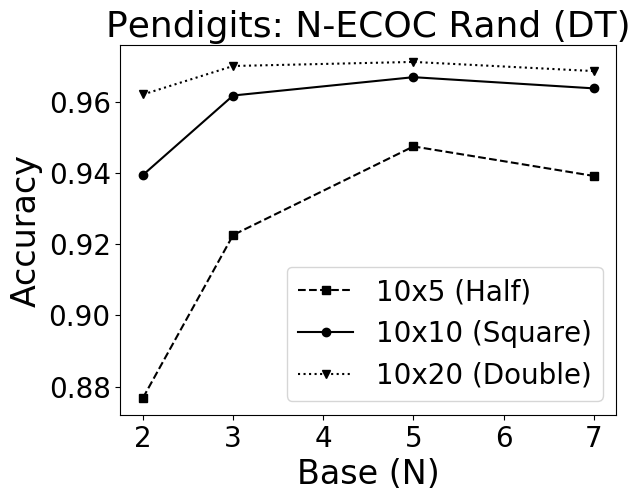}}
\caption{Random $N$-ECOC for Pendigits using Half, Square, and Double dimensions: Base vs. Accuracy (DT)}
\label{fig:pendigits-length-rand-DT}
\end{figure}

\begin{figure}
\centerline{\includegraphics[width=140pt,keepaspectratio]{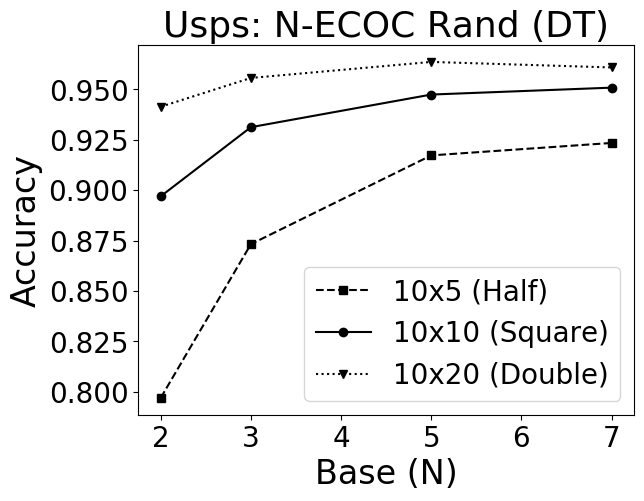}}
\caption{Random $N$-ECOC for Usps using Half, Square, and Double dimensions: Base vs. Accuracy (DT)}
\label{fig:usps-length-rand-DT}
\end{figure}

\begin{figure}
\centerline{\includegraphics[width=140pt,keepaspectratio]{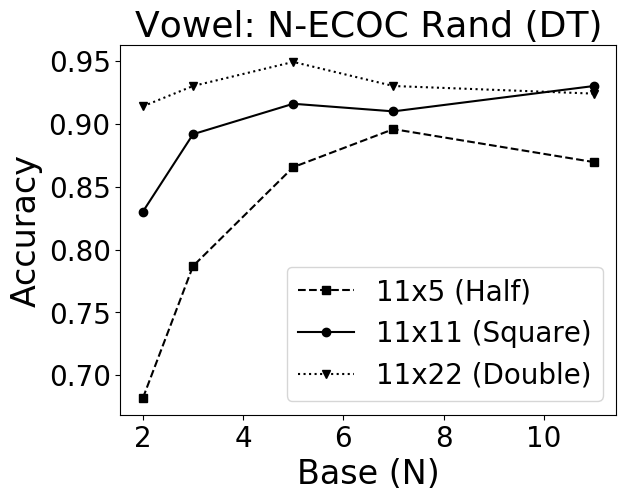}}
\caption{Random $N$-ECOC for Vowel using Half, Square, and Double dimensions: Base vs. Accuracy (DT)}
\label{fig:vowel-length-rand-DT}
\end{figure}

\begin{figure}
\centerline{\includegraphics[width=140pt,keepaspectratio]{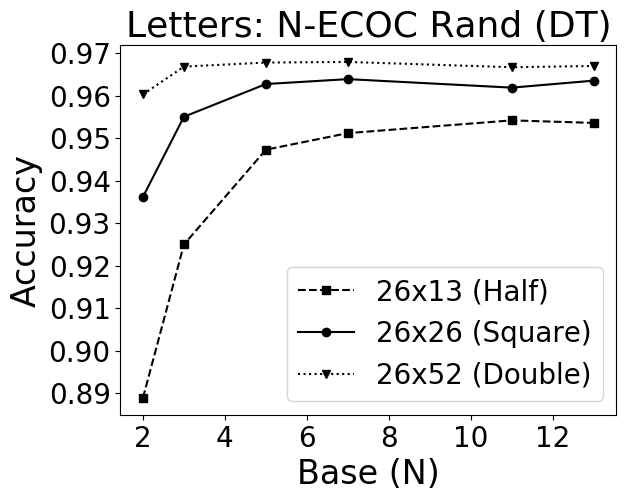}}
\caption{Random $N$-ECOC for Letters using Half, Square, and Double dimensions: Base vs. Accuracy (DT)}
\label{fig:letters-length-rand-DT}
\end{figure}

\begin{figure}
\centerline{\includegraphics[width=140pt,keepaspectratio]{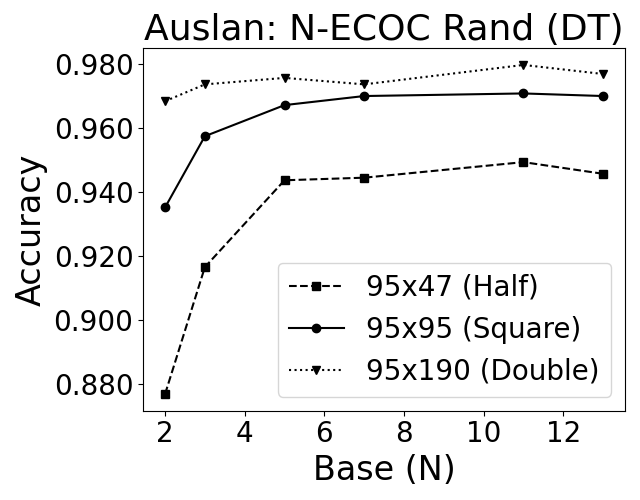}}
\caption{Random $N$-ECOC for Auslan using Half, Square, and Double dimensions: Base vs. Accuracy (DT)}
\label{fig:auslan-length-rand-DT}
\end{figure}

\begin{figure}
\centerline{\includegraphics[width=140pt,keepaspectratio]{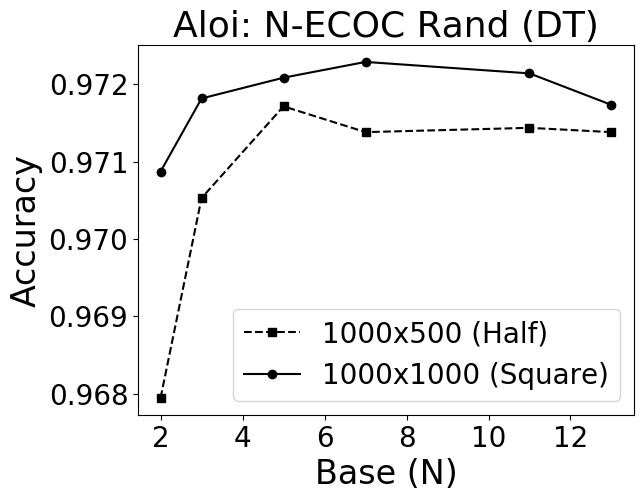}}
\caption{Random $N$-ECOC for Aloi using Half, Square, and Double dimensions: Base vs. Accuracy (DT)}
\label{fig:aloi-length-rand-DT}
\end{figure}

\noindent  3. Comparison of accuracy between half, square, and double for random $N$-ECOC using DT (Figures \ref{fig:pendigits-length-rand-DT}-\ref{fig:aloi-length-rand-DT})

\begin{figure}
\centerline{\includegraphics[width=140pt,keepaspectratio]{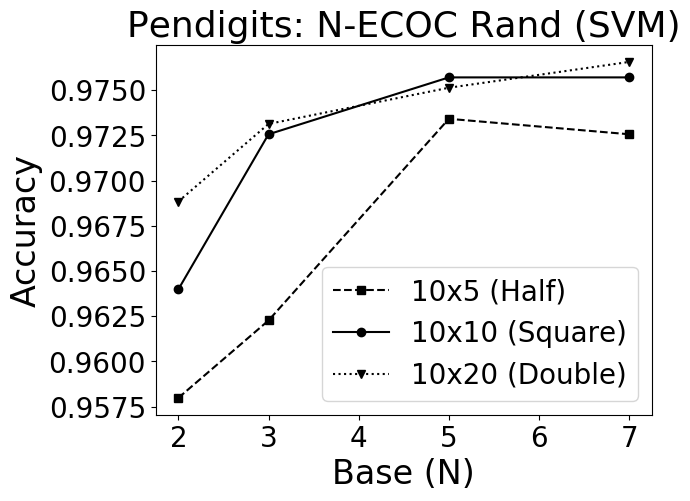}}
\caption{Random $N$-ECOC for Pendigits using Half, Square, and Double dimensions: Base vs. Accuracy (SVM)}
\label{fig:pendigits-length-rand-SVM}
\end{figure}

\begin{figure}
\centerline{\includegraphics[width=140pt,keepaspectratio]{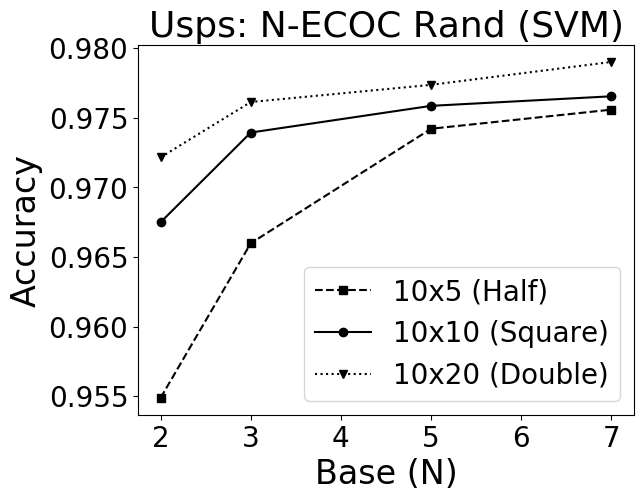}}
\caption{Random $N$-ECOC for Usps using Half, Square, and Double dimensions: Base vs. Accuracy (SVM)}
\label{fig:usps-length-rand-SVM}
\end{figure}

\begin{figure}
\centerline{\includegraphics[width=140pt,keepaspectratio]{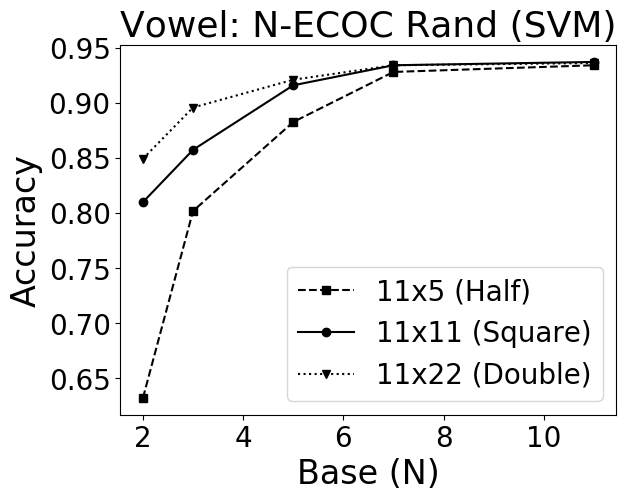}}
\caption{Random $N$-ECOC for Vowel using Half, Square, and Double dimensions: Base vs. Accuracy (SVM)}
\label{fig:vowel-length-rand-SVM}
\end{figure}

\begin{figure}
\centerline{\includegraphics[width=140pt,keepaspectratio]{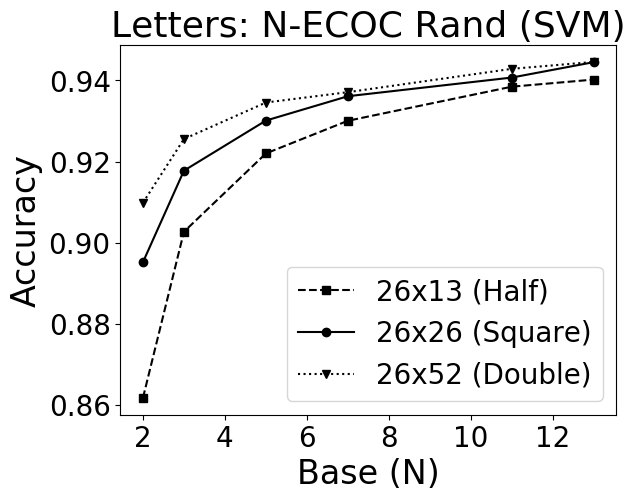}}
\caption{Random $N$-ECOC for Letters using Half, Square, and Double dimensions: Base vs. Accuracy (SVM)}
\label{fig:letters-length-rand-SVM}
\end{figure}

\begin{figure}
\centerline{\includegraphics[width=140pt,keepaspectratio]{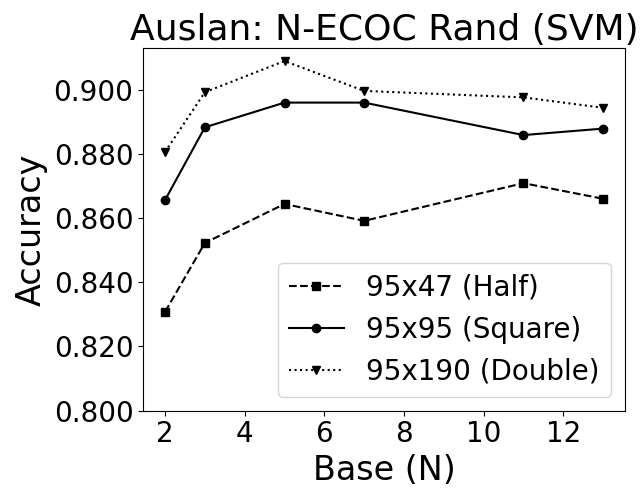}}
\caption{Random $N$-ECOC for Auslan using Half, Square, and Double dimensions: Base vs. Accuracy (SVM)}
\label{fig:auslan-length-rand-SVM}
\end{figure}

\begin{figure}
\centerline{\includegraphics[width=140pt,keepaspectratio]{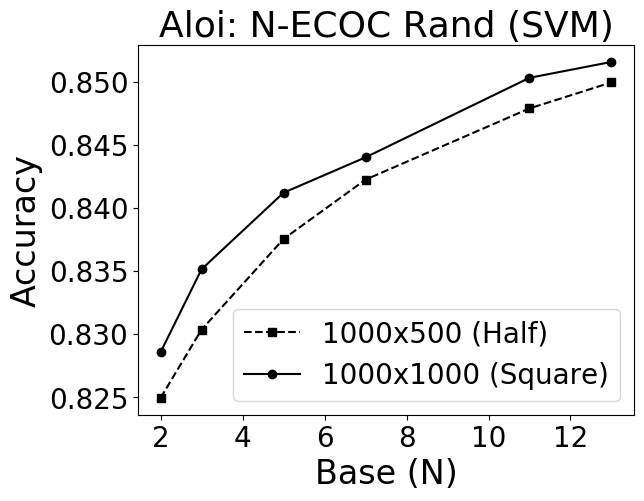}}
\caption{Random $N$-ECOC for Aloi using Half, Square, and Double dimensions: Base vs. Accuracy (SVM)}
\label{fig:aloi-length-rand-SVM}
\end{figure}

\noindent  4. Comparison of accuracy between half, square, and double for random $N$-ECOC using SVM (Figures \ref{fig:pendigits-length-rand-SVM}-\ref{fig:aloi-length-rand-SVM})

\end{document}